\documentclass[10pt,twocolumn,letterpaper]{article}

\usepackage{iccv}
\usepackage{times}
\usepackage{epsfig}
\usepackage{graphicx}
\usepackage{amsmath}
\usepackage{amssymb}

\usepackage{booktabs}
\usepackage{commath}
\usepackage{mathtools}
\usepackage{enumerate}
\usepackage{enumitem}
\usepackage{xcolor}
\usepackage{soul}

\usepackage[accsupp]{axessibility}

\iccvfinalcopy % *** Uncomment this line for the final submission

\def\iccvPaperID{****} % *** Enter the ICCV Paper ID here

\ificcvfinal\fi

\begin{document}

%%%%%%%%% TITLE
\title{Source-free Domain Adaptive Human Pose Estimation}

\author{Qucheng Peng, Ce Zheng, Chen Chen\\
Center for Research in Computer Vision,
University of Central Florida\\
{\tt\small \{qucheng.peng,ce.zheng\}@ucf.edu, chen.chen@crcv.ucf.edu}\\
%Center for Research in Computer Vision\\
%University of Central Florida\\
%{\tt\small chen.chen@crcv.ucf.edu}
}

\maketitle
\ificcvfinal\thispagestyle{empty}\fi

%%%%%%%%% ABSTRACT
\begin{abstract}

Human Pose Estimation (HPE) is widely used in various fields, including motion analysis, healthcare, and virtual reality. However, the great expenses of labeled real-world datasets present a significant challenge for HPE. To overcome this, one approach is to train HPE models on synthetic datasets and then perform domain adaptation (DA) on real-world data. Unfortunately, existing DA methods for HPE neglect data privacy and security by using both source and target data in the adaptation process.

To this end, we propose a new task, named source-free domain adaptive HPE, which aims to address the challenges of cross-domain learning of HPE without access to source data during the adaptation process. We further propose a novel framework that consists of three models: source model, intermediate model, and target model, which explores the task from both source-protect and target-relevant perspectives. The source-protect module preserves source information more effectively while resisting noise, and the target-relevant module reduces the sparsity of spatial representations by building a novel spatial probability space, and pose-specific contrastive learning and information maximization are proposed on the basis of this space. Comprehensive experiments on several domain adaptive HPE benchmarks show that the proposed method outperforms existing approaches by a considerable margin. \textcolor{magenta}{The codes are available at https://github.com/davidpengucf/SFDAHPE.}

\end{abstract}

%%%%%%%%% BODY TEXT
\section{Introduction}

%The task of 2D human pose estimation (HPE) from monocular images is crucial in computer vision, and deep neural networks have significantly improved their accuracy. However, like any other task, training deep learning models necessitates a vast amount of labeled data. Collecting real-world datasets for HPE, which must include diverse appearances and poses and generate ground truth annotations, is a time-consuming and labor-intensive process. On the other hand, synthetic datasets with annotations are cheaper and easier to obtain due to advancements in computer graphics and game engines. Therefore, one can generate a virtually unlimited number of labeled synthetic samples to train HPE models. Nevertheless, models trained on synthetic data may not perform well on real-world data due to the domain gap, or differences in data distribution. In such cases, domain adaptation (DA) is crucial to generalize synthetic data (source) trained models to real-world (target) data.

The accuracy of 2D human pose estimation (HPE) from monocular images has been significantly improved through the use of deep neural networks. However, obtaining sufficient labeled data for training such models is a laborious and time-consuming task, particularly for real-world datasets that require diverse appearances and poses with ground truth annotations. In contrast, the development of computer graphics and game engines has facilitated the generation of annotated synthetic datasets that can generate a virtually unlimited number of labeled samples. Nevertheless, models trained on synthetic data may not perform well on real-world data due to the domain gap or differences in data distribution. To address this challenge, domain adaptation (DA) is essential to enable synthetic data-trained models to generalize to real-world data.

Recently, source-free domain adaptation has gained significant attention due to its practicality in real-world scenarios, where using labeled source data during the adaptation is expensive or not feasible due to data privacy and security concerns. However, existing source-free domain adaptation methods in classification tasks may not be directly applicable to more complex tasks, such as 2D human pose estimation (HPE). This is because HPE requires estimating the spatial locations of different body joints, instead of just the object's category as classification does, making it more challenging to adapt from source to target domains. Therefore, there is a pressing need to develop effective source-free domain adaptation techniques specifically tailored to HPE tasks. This motivates us to propose a new task named source-free domain adaptive HPE in this paper.

The task of source-free domain adaptive HPE presents challenges from \textit{both the source-free DA and pose estimation perspectives}. Specifically, in traditional DA tasks, source data can participate in the adaptation process, allowing for a focus on reducing domain shifts between source and target. However, in source-free DA, only the pretrained source model can be used during adaptation, leading to catastrophic forgetting of the source domain. Directly applying the source pretrained model can also introduce noise due to domain shifts. Therefore, balancing the absorption of the source model's knowledge with the resistance of the source model's noise is critical. Current methods tend to adopt a progressive replacement strategy to alleviate the runoff of source information. For example, \cite{liang2020we,liang2022shot++} use the previous model to supervise ongoing adaptation, thereby evoking the source-related memory, while \cite{tarvainen2017mean,ge2020mmt} use the exponential moving average (EMA) strategy to slow down the forgetting of the source. Although these methods alleviate the negative impact of noise successfully, they modify the overall parameters of the pretrained source model significantly, leading to the disappearance of source-related knowledge and representations during the adaptation process. Therefore, it is necessary to develop source-protect modules that can protect source information more effectively while resisting noise at the same time. 

%Second, we present the challenges from a pose-specific perspective. Existing source-free DA methods are mainly based on the image classification task, which tends to enlarge the decision boundaries of different classes inside target domain. However, the regression space is continuous so it is impossible to find a clear decision boundary. Besides, all keypoints exist in a large discrete and sparse space, so it is not easy to apply methods that rely heavily on the matching and alignment of output probabilities. \cite{jiang2021regressive} observes that wrong predictions are more likely located at other keypoints, thus building a probability space based on the number of keypoints. Nevertheless, this probability is too small to represent all the potential locations where keypoints may appear, and this will limit the improvements of applying probability-based methods. In such a case, it is necessary to propose an effective method that represents the spatial distributions and deals with the task in a pose-specific way.

Moreover, we outline another challenge of source-free domain adaptive HPE: the sparsity of keypoints (i.e. key body joints) in an image. %\textcolor{blue}{Distribution alignment is a common technique route in DA, so building an ideal space for  spatial probability distributions is significant to our task. However, it is not easy to achieve on the basis of a complete image, because} 
When it comes to domain adaptation (DA) in classification, achieving distribution alignment to reduce domain gap is a straightforward process as the outputs can be represented in a probability space that includes all the categories.  However, the same cannot be said for Human Pose Estimation (HPE) based on a complete image, as each keypoint has a specific distance from others based on bone lengths, and the number of keypoints $K$ (e.g., 14 for human pose and 21 for hand pose) is significantly smaller than the number of pixels in an image (e.g., 512$\times$512$\approx$0.26M), making it a challenging task. Although heatmaps \cite{tompson2014joint,tompson2015efficient} are commonly used %to replace the complete image 
to predict keypoints (heatmap representation-based 2D HPE is considered as the mainstream approach for HPE) while reducing dimensionality in HPE, the number of pixels inside the heatmap is still considerable (e.g. 128$\times$128$\approx$16K) in comparison to $K$. Therefore, building a spatial probability space for the keypoints based on either the entire image or the heatmap is not ideal. \cite{jiang2021regressive} proposes constructing a low-dimensional distribution based solely on the $K$ keypoints, effectively transforming the HPE task into a $K$-classification task. However, this spatial distribution relies on clear decision boundaries between distinctive keypoints, which is not reasonable in HPE. %In other words, assuming either-or relations among various keypoints is not appropriate as mispredictions may occur at locations other than those of the other keypoints. 
In other words, all samples can be classified to a certain category in classification, but most pixels inside an image (e.g. more than 97$\%$ if the fault tolerance is decided by the metric PCK@0.05) can not be assigned to a certain keypoint in HPE, but mispredictions may occur at these unpresented locations. Hence, it is necessary to develop a novel spatial probability space that reduces sparsity while providing complete representations of the prediction space.

%To solve the aforementioned task-specific issues, we propose a novel framework that consists of the source model, intermediate model, and target model, exploring the source-free domain adaptive HPE task from both source-protect and pose-specific perspectives. In source-protect modules, the executions are conducted between the source model and the intermediate model to overcome catastrophic forgetting of source. We preserve the source-related information by transferring knowledge from the source model to the intermediate model while fixing the source model's feature extractor at the same time. In order to resist noise, we finetune on the source model's regressor, while applying a proposed residual loss to alleviate noise from the source model. In pose-specific modules, the executions are conducted between the intermediate model and the target model to alleviate domain shifts between source and target. We reduce the sparsity of spatial outputs by projecting them to two vectors in the directions of vertical and horizontal. In such a case, the reduction of dimensionality is assured while the completeness of spatial representations maintains. Based on the projected vectors, we  propose pose-specific contrastive learning and information maximization.  Our main contributions are summarized as follows:
To address the aforementioned task-specific challenges, we propose a novel framework that incorporates source-free domain adaptive HPE from both source-protect and target-relevant perspectives. Our framework consists of three models: the source model, the intermediate model, and the target model. Source model preserves source information, and target model absorbs knowledge from target data, while intermediate model interacts with both of them to reduce the domain gap between source and target implicitly, thereby used for the final inference. In the source-protect modules, we conduct knowledge transfer between the source model and the intermediate model to prevent catastrophic forgetting of the source-related information. We transfer knowledge from the source model to the intermediate model while keeping the source model's feature extractor fixed. To resist noise, we propose a new residual loss and finetune the source model's regressor. In the target-relevant modules, we conduct executions between the intermediate model and the target model to mitigate domain shifts between the source and target domains. We build a new spatial probability space for HPE to alleviate the sparsity problem. On the basis of this space, we propose contrastive learning and information maximization designed specifically for human pose estimation. Our contributions are summarized as follows:
\setlist{nolistsep}
\begin{itemize}[noitemsep,leftmargin=*] 
    \item We construct a new task named source-free domain adaptive human pose estimation, which focuses on the cross-domain learning of HPE without access to source data during the adaptation process.
    \item We propose a new framework consisting of three models: source model, intermediate model, and target model, which explores the task from both source-protect and target-relevant perspectives.
    \item We conduct comprehensive experiments on several domain adaptive HPE benchmarks and the results show that our method outperforms the existing approaches by a considerable margin. 
\end{itemize}
%------------------------------------------------------------------------
\section{Related Work}

%\textbf{Human Pose Estimation.} In this paper, we focus on domain adaptation in 2D HPE, which can be divided into regression methods and heatmap-based methods. Regression methods directly output keypoints coordinates. Starting from \cite{toshev2014deeppose}, deep neural networks are applied in HPE. \cite{sun2017compositional} introduces a structure-aware regression method that uses bone-based representations. \cite{li2021pose} combines vision transformers \cite{dosovitskiy2020image}, while \cite{shi2022end} proposes a fully end-to-end method trained in one single stage. For heatmap-based methods, the networks first obtain heatmaps constructed by putting 2D Gaussian kernels on the potential keypoints, then convert heatmaps to coordinates. \cite{chen2018cascaded} uses global and refine nets to improve performance. \cite{xiao2018simple} proposes Simple Baselines to simplify the network structures. \cite{sun2019deep} fully utilizes all the features at different scales. Besides, some work tends to improve the quality of heatmaps, like \cite{zhang2020distribution} that extends Gaussian to second order in heatmaps and \cite{luo2021rethinking} that proposes scale-adaptive heatmaps. \cite{wang2022regularizing} uses coupled embeddings to improve heatmap regression. In our model, we follow previous works \cite{jiang2021regressive,kim2022unified} to use the heatmap-based approach and apply Simple Baselines as the backbone. 

\textbf{Human Pose Estimation.} %2D human pose estimation (HPE) can be divided into regression methods and heatmap-based methods. Regression methods directly output keypoints coordinates, and have been widely used in HPE since the introduction of deep neural networks \cite{toshev2014deeppose}. Recent work has combined vision transformers \cite{dosovitskiy2020image} with regression methods \cite{li2021pose}, and proposed fully end-to-end methods trained in a single stage \cite{shi2022end}.
%Heatmap-based method, which is the mainstream technique route in 2D HPE, first obtains heatmaps constructed by putting 2D Gaussian kernels on the potential keypoints, and then convert these heatmaps to coordinates. \cite{chen2018cascaded} uses global and refine nets to improve performance, while \cite{xiao2018simple} proposes Simple Baselines to simplify the network structures. Other methods have fully utilized features at different scales \cite{sun2019deep}, or have aimed to improve the quality of heatmaps, such as by extending Gaussians to second order in heatmaps \cite{zhang2020distribution}, or proposing coupled embeddings to improve heatmap regression \cite{wang2022regularizing}. \emph{In this paper, we follow the protocols of other domain adaptive HPE works \cite{jiang2021regressive,kim2022unified} using heatmap-based HPE methods for source-free domain adaptation.}
The heatmap-based method \cite{shi2022end,wang2022regularizing} is currently a widely used technique in 2D HPE, which involves generating heatmaps by placing 2D Gaussian kernels on potential keypoints, followed by converting these heatmaps to coordinates. For instance, \cite{chen2018cascaded} uses global- and refine-nets, while \cite{xiao2018simple} simplifies the network structures through the Simple Baselines technique. Some approaches have focused on utilizing features at different scales \cite{sun2019deep}, while others aim to improve the quality of heatmaps by extending Gaussians to second order \cite{zhang2020distribution} or using coupled embeddings for heatmap regression \cite{wang2022regularizing}. \emph{In this study, we adopt the same heatmap-based HPE methods as other domain adaptive HPE works \cite{jiang2021regressive,kim2022unified}.}

%\textbf{Domain Adaptation.} Metric-based and GAN-based approaches are the two major routes in single-source domain adaptation. The metric-based methods measure the discrepancy between the source and target domain explicitly. \cite{long2015learning} uses maximum mean discrepancy, while \cite{tzeng2014deep} applies deep domain confusion. Recent work like \cite{deng2021cluster} jointly makes clustering and discrimination for alignment together with contrastive learning. The GAN-based methods originate from \cite{goodfellow2020generative} and build a min-max game for two players related to the source and target domains. \cite{ganin2015unsupervised} adopts domain to confuse the two players, while \cite{saito2018maximum} uses classifier discrepancy as the objective, and \cite{tang2020unsupervised} reveals that discriminative clustering on target will benefit the adaptation. However, the general domain adaptation setting needs access to source data, which raises concerns about data privacy and security, so Source-Free adaptation is proposed.

\textbf{General Domain Adaptation.} There are two main categories of methods for achieving general domain adaptation, which requires access to both the source and target data during adaptation. The first category is metric-based methods that explicitly measure the discrepancy between the source and target domains. For example, \cite{long2015learning} uses maximum mean discrepancy, and \cite{tzeng2014deep} applies deep domain confusion. Recent work, such as \cite{deng2021cluster}, jointly performs clustering and discrimination for alignment using contrastive learning \cite{chen2020simple,jiang2022supervised}. The second category is GAN-based \cite{goodfellow2020generative} methods that build a min-max game for two players related to the source and target domains. For example, \cite{ganin2015unsupervised,saito2018maximum} adopts domain confusion to confuse the two players, while \cite{tang2020unsupervised,pinyoanuntapong2023gaitsada} shows that discriminative clustering on the target benefits adaptation. \emph{However, access to source data is required in the general domain adaptation setting, which raises concerns about data privacy and security. }

%\textbf{Source-Free Domain Adaptation.} Based on the work of \cite{li2020ma,liang2020we}, Source-Free has become the mainstream paradigm for alleviating concerns about data privacy and security in domain adaptation. There are two technique routes under the source-free setting: self-supervision and virtual source transfer. For the self-supervised methods, \cite{liang2020we} is the most representative one, which introduces information maximization to assist adaptation. \cite{xia2021adaptive} treats the problem from a discriminative perspective and adds a specific representation learning module to help the generalization, and \cite{chen2022contrastive} proposes the online pseudo label refinement. As for virtual source methods, most of them build GANs to generate virtual source data. \cite{kurmi2021domain} uses conditional GAN to generate new samples, while \cite{hou2021visualizing} provides interesting visualizations for unseen knowledge and \cite{li2020ma} applies collaborative GAN to achieve better generations. But most source-free domain adaptation methods for classification are not applicable to HPE tasks, and that's the reason why we propose the new topic source-free domain adaptive HPE.

\textbf{Source-Free Domain Adaptation.} Source-Free has become the mainstream paradigm for alleviating concerns about data privacy and security in domain adaptation, as shown in the work of \cite{li2020ma,liang2020we,peng2022toward}. There are two technique routes under the source-free setting: self-supervision and virtual source transfer. For the self-supervised methods, \cite{liang2020we} is the most representative, which introduces information maximization to assist adaptation. Other self-supervised methods include \cite{xia2021adaptive}, which treats the problem from a discriminative perspective, and \cite{chen2022contrastive,peng2022toward}, which proposes online pseudo label refinement. In virtual source methods, most build GANs to generate virtual source data. Examples include conditional GANs used in \cite{kurmi2021domain}, and collaborative GANs in \cite{li2020ma}, which achieve better generations. \emph{However, most source-free DA methods for classification are not applicable to HPE tasks due to the sparsity of keypoints' distributions.}

%\textbf{Domain Adaptive Human Pose Estimation.} Domain-adaptive HPE methods can be divided into two categories. One is the shared structure, in which the networks of source pretrain and target adaptation share weights. CC-SSL \cite{mu2020learning} uses one network trained in an end-to-end fashion. RegDA \cite{jiang2021regressive} applies one shared feature extractor and two separate regressors. TransPar \cite{han2022transpar} adopts a similar structure but focuses on transferable parameters. The other adopts an unshared structure, where the pretrain and adaptation form a teacher-student paradigm. To name a few, MDAM \cite{li2021synthetic} tackles this structure, together with a novel pseudo-label strategy. UniFrame \cite{kim2022unified} modifies the classic Mean-Teacher \cite{tarvainen2017mean} model, combining with style transfer \cite{huang2017arbitrary}. MarsDA \cite{jin2022branch} edits RegDA in a teacher-student manner. However, all these HPE methods need access to source data, which raises concerns about data privacy and security, so Source-Free adaptation for HPE is necessary.

\textbf{Domain Adaptive Human Pose Estimation.} There are two main categories of methods used for domain-adaptive HPE. The first category involves a shared structure, where the weights of the networks from both the source pretrain and target adaptation are shared. CC-SSL \cite{mu2020learning} employs a single end-to-end trained network, RegDA \cite{jiang2021regressive} utilizes one shared feature extractor and two separate regressors, and TransPar \cite{han2022transpar} emphasizes transferable parameters using a similar structure. The second category is the unshared structure, where the pretrain and adaptation create a teacher-student paradigm. In this category, MDAM \cite{li2021synthetic} addresses this structure alongside a novel pseudo-label strategy, UniFrame \cite{kim2022unified} modifies the classic Mean-Teacher \cite{tarvainen2017mean} model by combining it with style transfer \cite{huang2017arbitrary}, and MarsDA \cite{jin2022branch} applies a teacher-student approach to edit RegDA.
%There are two categories of methods for domain-adaptive HPE. The first category is the shared structure, where the networks of source pretrain and target adaptation share weights. CC-SSL \cite{mu2020learning} uses a single end-to-end trained network, RegDA \cite{jiang2021regressive} applies one shared feature extractor and two separate regressors, and TransPar \cite{han2022transpar} focuses on transferable parameters using a similar structure. The second category is the unshared structure, where the pretrain and adaptation form a teacher-student paradigm. MDAM \cite{li2021synthetic} tackles this structure along with a novel pseudo-label strategy, UniFrame \cite{kim2022unified} modifies the classic Mean-Teacher \cite{tarvainen2017mean} model by combining it with style transfer \cite{huang2017arbitrary}, and MarsDA \cite{jin2022branch} uses a teacher-student approach to edit RegDA. %\emph{However, all these HPE methods require access to source data, which raises concerns about data privacy and security. Therefore, Source-Free adaptation for HPE is necessary.}
\emph{However, \textbf{all these existing domain adaptive HPE approaches are not source-free}, which raises concerns about data privacy and security. Therefore, it is necessary to consider source-free domain adaptive HPE. To our knowledge, we are the first work to study this challenging problem.}

%In summary, source-free domain adaptation is a necessary paradigm due to data privacy and security concerns. However, the existing source-free domain adaptation methods that are designed for classification tasks may not be directly applicable to the HPE task. Therefore, we propose a new task named source-free domain adaptive HPE.

\section{Method}

\subsection{Preliminary of 2D HPE}
\label{sec:pre}
%\vspace{-5pt}
\begin{figure}[!ht]
  \centering
   \includegraphics[width=1.0\linewidth]{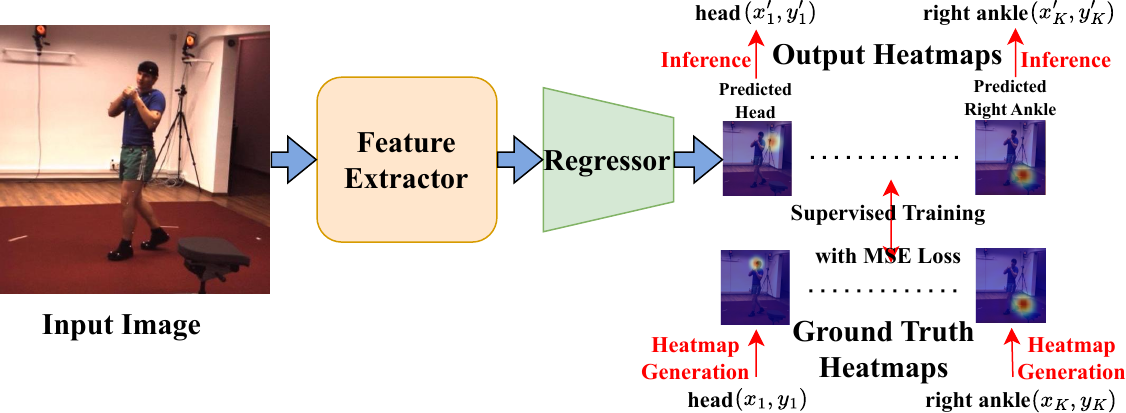}

   \caption{Process of heatmap-based 2D HPE. The model includes two components: the feature extractor $G$ and the regressor $F$. After feeding the input image to the model, we obtain $K$ output heatmaps corresponding to $K$ keypoints.
   }%\vspace{-7pt}
   \label{fig:hpe}
\end{figure}
%In 2D HPE, we have a labeled pose dataset $\mathcal{D} = \{(x_{i},y_{i})\}$, where $x_{i} \in \mathbb{R}^{C \times H \times W}$ is the image and $y_{i} \in \mathbb{R}^{K\times 2}$ is the corresponding keypoint coordinates. Here $H$ and $W$ are the height and width of the image and $C$ is the number of channels. Besides, $K$ is the number of keypoints. For the sake of more smooth training, most existing 2D methods use heatmaps $HM_{i} \in \mathbb{R}^{K \times H' \times W'}$ to represent coordinates in a spatial way during supervised learning, where $H'$ and $W'$ are the height and width of heatmaps. In the inference stage, well-trained supervised models output heatmaps that predict the probability of joints occurring at each pixel. For each heatmap, the pixel with the highest probability is treated as the predicted joint's location. 

\begin{figure*}[t]
%\vspace{-5pt}
  \centering
   \includegraphics[width=1.00\linewidth]{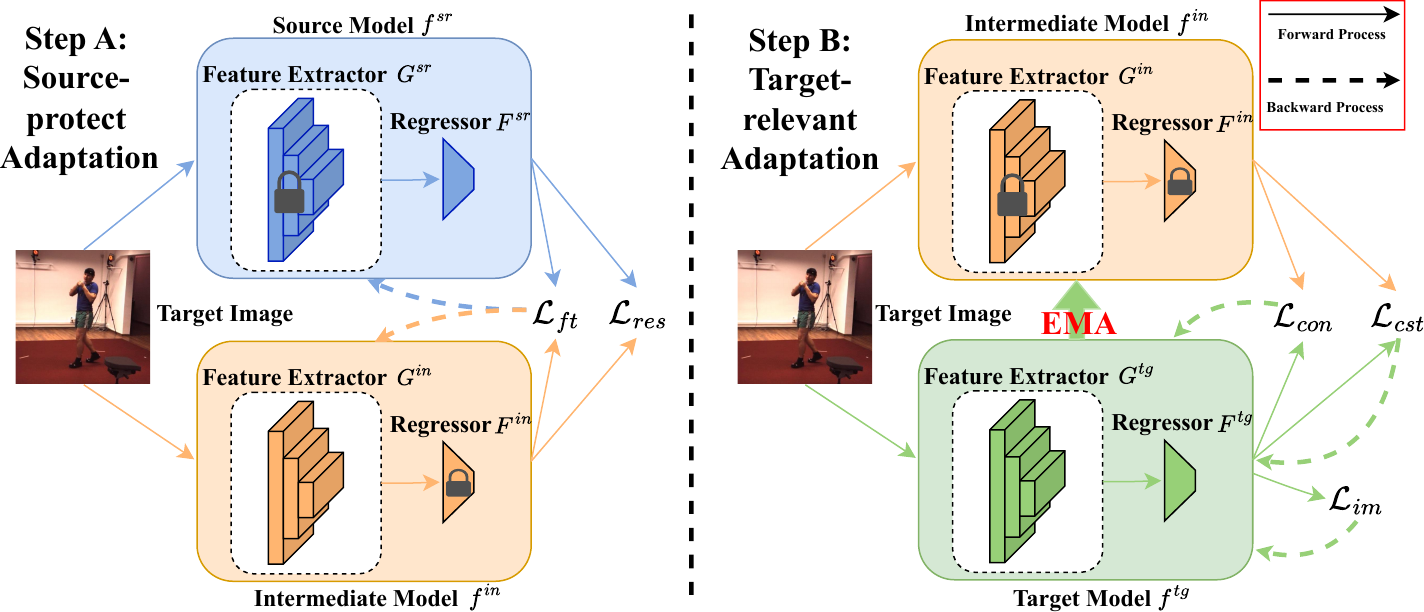}%\vspace{-1mm}
   %\vspace{-7pt}
   \caption{Overall framework of our method. It includes three models, each with one feature extractor and one regressor. The adaptation contains two steps: Step A and Step B. Step A is applied for source-protect adaptation, while Step B aims at target-relevant adaptation. The solid arrows show the forward process of associated losses, and the dotted arrows show the backward updating process of associated losses. EMA means the exponential moving average process from target model to intermediate model to update intermediate model's weights.
   }%\vspace{-11pt}
   \label{fig:framework}
\end{figure*}
In 2D HPE, we are given a labeled pose dataset $\mathcal{D} = {(x_i,y_i)}$, where $x_i \in \mathbb{R}^{C \times H \times W}$ is an image and $y_i \in \mathbb{R}^{K \times 2}$ are the corresponding keypoint coordinates. Here, $H$ and $W$ represent the height and width of the image, respectively, and $C$ is the number of channels. Additionally, $K$ represents the number of keypoints. To ensure smoother training, most existing 2D methods represent the keypoint coordinates as heatmaps $HM_i \in \mathbb{R}^{K \times H' \times W'}$, where $H'$ and $W'$ are the height and width of the heatmaps, respectively. These heatmaps are used to represent the coordinates in a spatial way. The process of how the model works is defined in Fig. \ref{fig:hpe}. Ground truth heatmaps are generated by a 2D Gaussian centered at the ground truth joint location, with one heatmap for each keypoint. We denote the transform from coordinates to heatmaps as $T:  \mathbb{R}^{K\times 2} \rightarrow \mathbb{R}^{K \times H'\times W'}$. %In this figure,  %Moreover, any single model in this paper consists of two parts: the feature extractor $G$ and the regressor $F$. 
The training processes are conducted between the output heatmaps and ground truth heatmaps using MSE for supervised learning. In the inference stage, well-trained supervised models output heatmaps that predict the probability of joints occurring at each pixel. For each heatmap, the pixel with the highest probability is considered to be the predicted location of the joint.%, which can be represented as  $T^{-1}:\rightarrow \mathbb{R}^{K \times H'\times W'} \rightarrow  \mathbb{R}^{K\times 2}$.

\subsection{Problem Statement}

In the source-free domain adaptive HPE setting, we have a source domain dataset $\mathcal{S} = \{(x^{s}_{i},y^{s}_{i})\}^{n_{s}}_{i=1}$ with $n_{s}$ labeled pose samples. $x^{s}_{i} \in \mathbb{R}^{C \times H \times W}$ is the source image and $y^{s}_{i} \in \mathbb{R}^{K\times 2}$ is the corresponding pose annotation. Besides, there exists a target domain dataset $\mathcal{T} = \{x^{t}_{i}\}^{n_{t}}_{i=1}$ that includes $n_{t}$ unlabeled pose samples. The source and target domains share the same label space but lie in different distributions (e.g. synthetic vs. real). The training procedure is split into two stages: the pretrain stage and the adaptation stage. In the pretrain stage, the labeled source domain dataset $\mathcal{S}$ is used to train the source model $f^{sr}$. \ul{However, what distinguishes source-free domain adaptive HPE from general domain adaptive HPE is the adaptation stage, where only the source model $f^{sr}$ and target domain dataset $\mathcal{T}$ are available for training, and the source domain dataset cannot be used anymore.} Our objective is to develop a model that performs well on the target dataset.%Our goal is to use the labeled source dataset and unlabeled target dataset to obtain a model that achieves high performance on the target dataset.

%In the setting of source-free domain adaptive HPE, the conditions are even more challenging. The training procedure is split into two stages: the pretrain stage and the adaptation stage. In the pretrain stage, the labeled source domain dataset $\mathcal{S}$ is used to train the source model $f^{sr}$. However, what distinguishes source-free domain adaptive HPE from general domain adaptive HPE is the adaptation stage, where only the source model $f^{sr}$ and target domain dataset $\mathcal{T}$ are available for training, and the source domain dataset cannot be used anymore. Our objective is still to develop a model that performs well on the target dataset.

\subsection{Pipeline of Our Method}

\subsubsection{Source Model Pretrain}

For DA in HPE, labeled source data are applied for pretraining. Same as most existing domain adaptive pose estimation methods \cite{mu2020learning,li2021synthetic,jiang2021regressive,kim2022unified}, the network is composed of a feature extractor $G$ (such as ResNet backbone) and a pose regressor $F$, so the source model can be represented as $f^{sr} = F^{sr}(G^{sr}(\cdot))$. Based on the supervised heatmap-based loss (MSE loss), we propose the overall objective for source pretraining:
\begin{equation}
\begin{multlined}
    \mathcal{L}_{pretrain} = \mathbb{E}_{(x^{s}_{i},y^{s}_{i}) \in \mathcal{S}} \mathcal{L}_{mse}(F^{sr}(G^{sr}(x_{i}^{s})),T(y_{i}^{s})),
\end{multlined}
\label{eq:pretrain}
\end{equation}where $\mathcal{L}_{mse}$ is the Mean Squared Error (MSE) loss between the heatmap representations of prediction and ground truth, and $T$ is the transform from coordinates to heatmaps defined in Sec. \ref{sec:pre}.

%\vspace{-10pt}
\subsubsection{Adaptation Framework}

As Fig. \ref{fig:framework} shows, our method contains three models: source model $f^{sr}=F^{sr}(G^{sr}(\cdot))$, intermediate model $f^{in}=F^{in}(G^{in}(\cdot))$, and target model $f^{tg}=F^{tg}(G^{tg}(\cdot))$. Each model includes one feature extractor and one regressor. Assume the heatmaps generated by source model as \begin{small}$\boldsymbol{H_{i}^{sr}}=F^{sr}(G^{sr}(x_{i}^{t}))= \begin{bmatrix}
        h_{i,1}^{sr}, h_{i,2}^{sr},...,h_{i,K}^{sr}
\end{bmatrix}^{\top}$\end{small}, and those from intermediate model and target model respectively as \begin{small}$\boldsymbol{H_{i}^{in}}=F^{in}(G^{in}(x_{i}^{t}))= \begin{bmatrix}
        h_{i,1}^{in}, h_{i,2}^{in},...,h_{i,K}^{in}
\end{bmatrix}^{\top}$\end{small} and \begin{small}$\boldsymbol{H_{i}^{tg}}=F^{tg}(G^{tg}(x_{i}^{t}))= \begin{bmatrix}
        h_{i,1}^{tg}, h_{i,2}^{tg},...,h_{i,K}^{tg}
\end{bmatrix}^{\top}$\end{small}, where $K$ is the number of keypoints. We also assume the keypoint indexing space as \begin{small}$\mathcal{P} = [1,2,...,K]$\end{small}. The source model is used to maintain source information, while the target model is applied to learn target knowledge. The intermediate model is the transition model between these two models. It serves as a bridge between source and target as it learns from source and target simultaneously, thus eliminating the domain gap implicitly. Therefore, it outperforms both the source and target models and is used for the final inference. When starting the adaptation, $f^{in}$ and $f^{tg}$ are initialized from the source model $f^{sr}$.  For a single adaptation iteration, two steps are executed as shown in Fig. \ref{fig:framework}. Step A is for the source-protect adaptation, and Step B aims to adapt in a target-relevant way.

In Step A, our goal is to overcome catastrophic forgetting on source domain, while eliminating the noise from the source model due to domain shift. Therefore, the source model's feature extractor $G^{sr}$ is fixed to retain source knowledge, while its regressor $F^{sr}$ keeps updating to improve regression accuracy, hence reducing noises in the outputs. On the opposite, the intermediate model's regressor $F^{in}$ is fixed, but its feature extractor $G^{in}$ continues to update. That's because we want the feature extractor to learn more source knowledge on the representation level, while the noise from the source model due to domain shifts can be inhibited.
%the accuracy estimation can be ensured. 

In Step B, we aim to conduct the adaptation between the intermediate model and the target model in a target-relevant manner. During the process, several pose-specific techniques are utilized to minimize the discrepancy between these two models. Though both $f^{in}$ and $f^{tg}$ are updated at this step, $f^{tg}$ is updated via back-propagation while $f^{in}$ is via exponential moving average from $f^{tg}$, since averaging model weights over training steps tends to produce a more accurate model than using the final weights directly \cite{tarvainen2017mean}.  

In the following two subsections, we introduce the detailed techniques used for Step A and Step B. 
%\vspace{-5pt}
\subsubsection{Source-protect Modules}

In this part, we propose modules related to source-free-specific adaptation. The biggest challenge is to distill knowledge from the source model and resist the potential noise due to domain shift. First, we introduce the finetune loss that calibrates on the source model:
%\vspace{-5pt}
\begin{equation}
\begin{multlined}
    \mathcal{L}_{ft} = \mathbb{E}_{x^{t}_{i} \in \mathcal{T}} \mathcal{L}_{mse}(F^{sr}(G^{sr}(x_{i}^{t})), F^{in}(G^{in}(x_{i}^{t}))). 
\end{multlined}
\label{eq:finetune}
%\vspace{-5pt}
\end{equation} To be notified, we hope the regression results of the source model can be more accurate, while the source knowledge can be retained. Thus, the feature extractor is fixed so that the source representations continue to be generated, and the regressor is being updated:
%\vspace{-8pt}
\begin{equation}
    \theta_{F^{sr}}^{t+1} = \theta_{F^{sr}}^{t} - \lambda^{sr}\frac{\partial L_{ft}}{\partial \theta_{F^{sr}}^{t}},
\label{eq:finetune-update}
%\vspace{-6pt}
\end{equation}where $\theta_{F}^{s}$ is the weights of the source model's regressor $F^{s}$, while $t$ denotes the training step and $\lambda^{s}$ is the source model's learning rate. However, improving source model is not enough. What's more, we propose the residual loss to improve intermediate model.

\begin{figure}[t]
%\vspace{-4pt}
  \centering
   \includegraphics[width=1.0\linewidth]{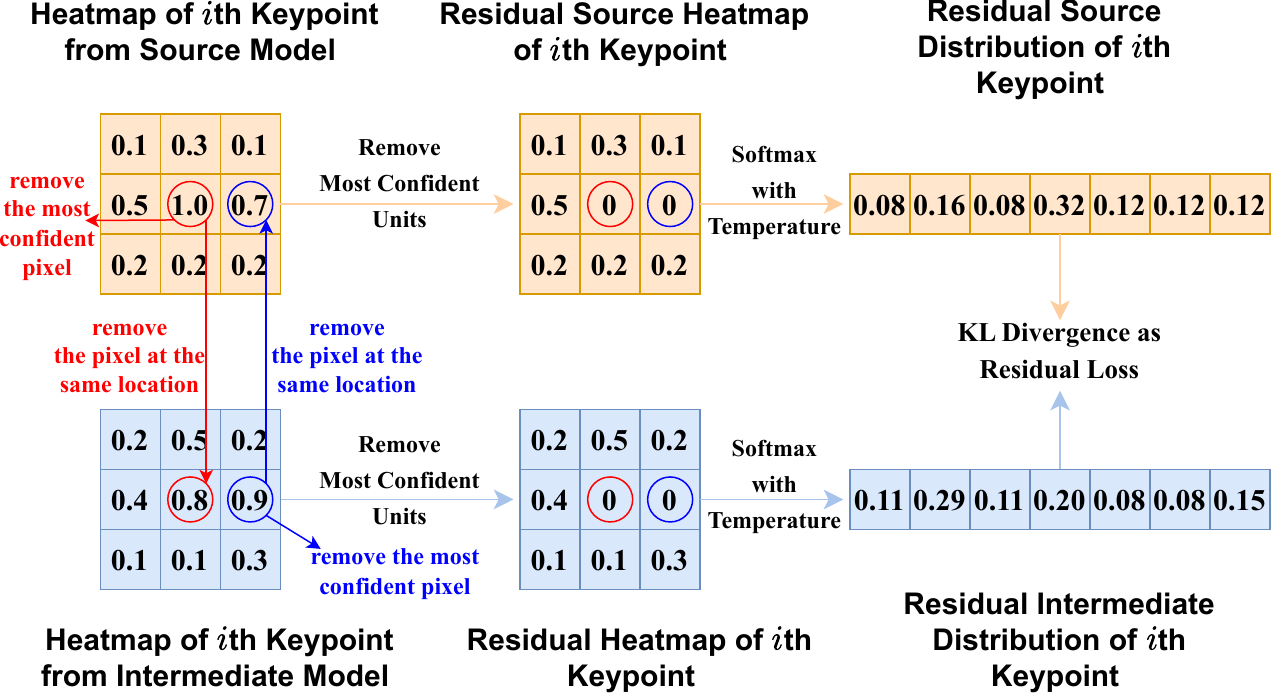}%\vspace{-1mm}
   \caption{Proposed residual loss. Here we choose two heatmaps at the same location generated from source model and intermediate model for illustration. For simplicity, here the heatmap's scale is set as $3\times 3$, while the actual size is $64\times64$. 
   }%\vspace{-10pt}
   \label{fig:residual}
\end{figure}

The application of Equation \ref{eq:finetune} facilitates mutual knowledge transfer between the source and intermediate models. Apart from the pixel with the highest confidence in a heatmap, the other pixels also contain valuable information that benefits the knowledge transfer, and that's why we propose residual loss in Fig. \ref{fig:residual}. From the example in the figure, the most confident pixel (confidence = 1.0) in the source heatmap is removed, along with the pixel  at the same location which has a confidence of 0.8 in the intermediate heatmap marked with red. The pair marked with blue follows the same rule but on the basis of intermediate heatmap. Then we build adaptive residual heatmaps
%However, the presence of highly confident pixels in the heatmaps inhibits the representations of other pixels, particularly after the softmax operation, resulting in the destruction of source knowledge during transfer to the intermediate model. To overcome this limitation, we propose a residual loss as illustrated in Fig. \ref{fig:residual}. We remove the most confident points from both heatmaps at the $j$th keypoint to obtain residual heatmaps, 
denoted as $\hat{h}_{i,j}^{sr}$ and $\hat{h}_{i,j}^{in}$. Moreover, we employ KL divergence to ensure that $\hat{h}_{i,j}^{in}$ approaches $\hat{h}_{i,j}^{sr}$, thereby preserving the source information. Based on these, we deduce the residual loss as:
%\vspace{-3pt}
\begin{equation}
    \mathcal{L}_{res} = \mathbb{E}_{x^{t}_{i} \in \mathcal{T}} \mathbb{E}_{j \in \mathcal{P} }\mathrm{D_{KL}} (\sigma(\hat{h}_{i,j}^{sr} / \tau)||\sigma(\hat{h}_{i,j}^{in} / \tau)), 
\label{eq:res}
%\vspace{-2pt}
\end{equation}where $\mathrm{D_{KL}}$ is the KL divergence, and $\sigma(\cdot)$ is the softmax function. Besides, $\tau$ is the temperature used to scale the residual heatmaps and is empirically set to $0.3$. By combining the finetune loss and the residual loss, we get the objective function with the optimization process for updating the intermediate model:
%\vspace{-7pt}
\begin{equation}
     \mathcal{L}_{in} = \mathcal{L}_{ft} + \alpha \mathcal{L}_{res}, ~~~
    \theta_{G^{in}}^{t+1} = \theta_{G^{in}}^{t} - \lambda^{in}\frac{\partial L_{in}}{\partial \theta_{G^{in}}^{t}},
\label{eq:in}
%\vspace{-5pt}
\end{equation} where $t$ is the training step, and $\lambda^{in}$ is the learning rate for $G^{in}$. Moreover, $\alpha$ is the trade-off hyperparameter between these two losses.

%\vspace{-5pt}
\subsubsection{Target-relevant Modules}

\begin{figure}[t]
%\vspace{-5pt}
  \centering
  \includegraphics[width=1.0\linewidth]{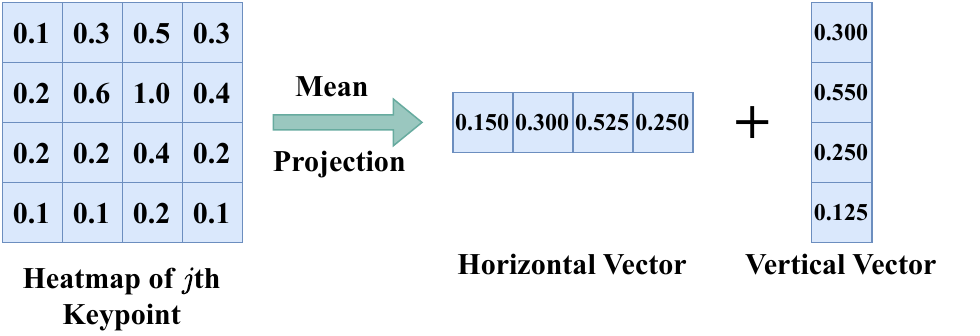}%\vspace{-1mm}
  \caption{Illustration of the projection of a heatmap. In order to reduce the sparsity of a heatmap, we project it into a horizontal vector and a vertical vector. For simplicity, here the heatmap's scale is set as $4\times 4$ and the vector's scale as $4\times 1$, while the actual heatmap's size is $64\times64$ and the vector's size is $64\times 1$.  
   }%\vspace{-5pt}
   \label{fig:spa}
\end{figure}

This section introduces a module that is associated with target information. As previously mentioned, the sparsity of keypoints in the image presents a significant challenge in our task. To address this issue, we propose using projected horizontal and vertical vectors of heatmaps as spatial representations, as shown in Fig. \ref{fig:spa}. This approach reduces the sparsity of the spatial probability space effectively since the number of overall digits inside the heatmap is $64\times64 \approx 4K$, while the number of units in each vector is 64. Therefore, we can effectively reduce the sparsity of the spatial probability space. With 64 units in each vector, the spatial probability space size is acceptable compared to the number of keypoints.

Despite the reduction in sparsity, the completeness of spatial representations is not affected. That is because all the coordinates can always be divided into two representations, horizontally and vertically. In such a case, our spatial representation is better than the $K$-classification probability space proposed by \cite{jiang2021regressive}, which only includes the keypoint-relevant information and does not take the units in the heatmap that are not keypoints into consideration.

In Fig. \ref{fig:spa}, we introduce the projection of a heatmap from both horizontal and vertical directions. For the heatmap $h_{i,j}^{in}$ generated by the intermediate model from $i$th sample's $j$th keypoint, its projection can be represented as $proj(h_{i,j}^{in})=(v_{i,jx}^{in}, v_{i,jy}^{in})$, where $v_{i,jx}^{in}$ is the horizontal vector and $v_{i,jy}^{in}$ is the vertical vector. This vector pair is the foundation of the following modules and losses that we propose.

\begin{figure}[t]
%\vspace{-5pt}
  \centering
   \includegraphics[width=1.0\linewidth]{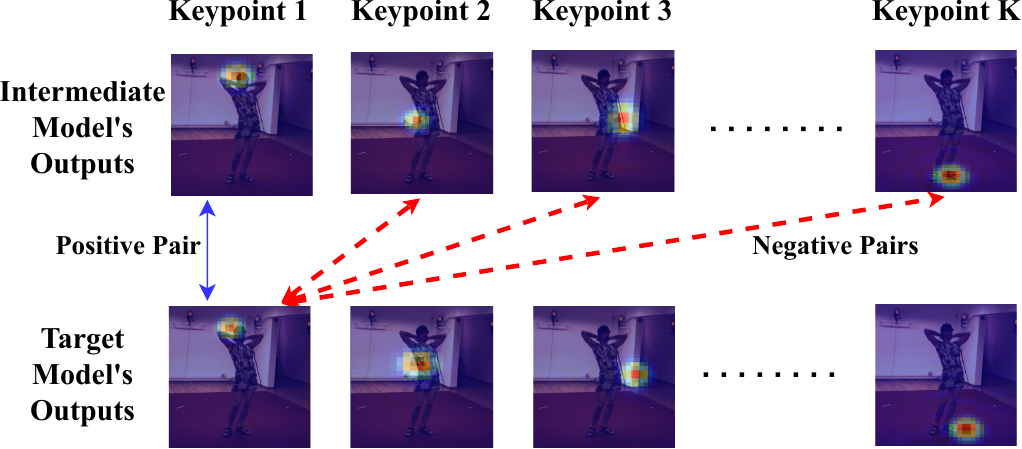}%\vspace{-1mm}
   \caption{Illustration of building positive and negative keypoint pairs for contrastive learning between the output heatmaps of the intermediate model and target model.
   }%\vspace{-10pt}
   \label{fig:cst}
\end{figure}

First, we propose the pose-specific contrastive loss between the intermediate model and target model. According to the observation from \cite{jiang2021regressive}, wrong predictions are usually located at other keypoints' positions. In such a case, improving regression can be achieved by minimizing the discrepancy of heatmaps at the same location and maximizing the discrepancy of heatmaps at different locations. And that's how we define the positive and negative pair in Fig. \ref{fig:cst}. Denote the similarity of two heatmaps generated by the intermediate model and target model respectively as:
%\vspace{-5pt}
\begin{small}
\begin{equation}
sim(h_{i,j}^{in}, h_{i,k}^{tg}) = \frac{1}{2}(\frac{v_{i,jx}^{in} \cdot v_{i,kx}^{tg}}{||v_{i,jx}^{in}||||v_{i,kx}^{tg}||}+\frac{v_{i,jy}^{in} \cdot v_{i,ky}^{tg}}{||v_{i,jy}^{in}||||v_{i,ky}^{tg}||}).
\label{eq:sim}
\end{equation}%\vspace{-8pt}
\end{small}On the basis of the similarity, we define the overall pose-specific contrastive loss as:
%\vspace{-5pt}
\begin{small}
\begin{equation}
\mathcal{L}_{cst} = -\mathbb{E}_{x^{t}_{i} \in \mathcal{T}} \mathbb{E}_{j \in \mathcal{P}} \log \frac{\exp(sim(h_{i,j}^{in}, h_{i,j}^{tg}))}{\Sigma_{k \in \mathcal{P}}\exp(sim(h_{i,j}^{in}, h_{i,k}^{tg}))}.
\label{eq:cst}
\end{equation}
\end{small}

Next, we propose a pose-specific self-supervised technique called pose-specific information maximization. Information maximization is very helpful to source-free domain adaptation for classification \cite{liang2020we} because it makes the target
outputs individually certain and globally diverse, hence mitigating the domain gap implicitly. However, the improvements are limited due to the sparsity of heatmaps. With the help of projected vectors, we refine the information maximization loss as:
%\vspace{-5pt}
\begin{equation}
    \label{eq:im-1}
    %\small
    \mathcal{L}_{im} = \mathbb{E}_{x^{t}_{i} \in \mathcal{T}} (\mathcal{L}_{entx}(x^{t}_{i}) + \mathcal{L}_{enty}(x^{t}_{i}) - \mathcal{L}_{div}(x^{t}_{i})), \\
\end{equation}
%\vspace{-10pt}
\begin{equation}
\label{eq:im-2}
%\small
\left\{\begin{aligned}
    \mathcal{L}_{entx}(x^{t}_{i}) &= -\mathbb{E}_{j \in \mathcal{P}} [\sigma(v_{i,jx}^{tg})\log \sigma(v_{i,jx}^{tg})], 
    \\
    \mathcal{L}_{enty}(x^{t}_{i}) &= -\mathbb{E}_{j \in \mathcal{P}} [\sigma(v_{i,jy}^{tg})\log \sigma(v_{i,jy}^{tg})], \\
    \mathcal{L}_{div}(x^{t}_{i}) &= -[\mathbb{E}_{j \in \mathcal{P}} \sigma(h_{i,j}^{tg})]\log [\mathbb{E}_{j \in \mathcal{P}} \sigma(h_{i,j}^{tg})].
\end{aligned} \right.
\end{equation} Here $\mathcal{L}_{entx}$ is the entropy of horizontal vectors, and $\mathcal{L}_{enty}$ is the entropy of vertical vectors. The last term in Equation \ref{eq:im-1} is the diversity-promoting objective. Moreover, we introduce the consistency loss between the two models' outputs:
\begin{small}
\begin{equation}
\mathcal{L}_{con} = \mathbb{E}_{(x^{s}_{i},y^{s}_{i}) \in \mathcal{S}} \mathcal{L}_{mse}(F^{in}(G^{in}(x_{i}^{s})),F^{tg}(G^{tg}(x_{i}^{s}))).
\label{eq:con}
\end{equation}
\end{small}

By summarizing above, we conclude the overall objective loss function for the target model as:
%\vspace{-5pt}
\begin{equation}
\label{eq:tgt}
    \mathcal{L}_{tgt} =  \mathcal{L}_{con} + \beta \mathcal{L}_{cst} + \gamma \mathcal{L}_{im},
\end{equation}
 where $\beta$ and $\gamma$ are trade-off parameters.

Moreover, the intermediate model is updated via the exponential moving average strategy (EMA) in this step:
%\vspace{-5pt}
\begin{equation}
\label{eq:ema}
    \theta^{t+1}_{f^{in}} = \eta \theta^{t}_{f^{in}} + (1 - \eta)\theta^{t}_{f^{tg}}, 
\end{equation}
where $t$ denotes the step of training and $\eta$ denotes the smoothing coefficient set to 0.999 by default.

\section{Experiments}

\textbf{\textcolor{blue}{Datasets.}} We use three human pose datasets and three hand pose datasets to validate our approach. \textbf{SURREAL} \cite{varol2017learning} is utilized as the source dataset and encompasses six million synthetic human pose images. \textbf{Human3.6M} \cite{ionescu2013human3}, one of the target human pose datasets, is a frequently used real-world dataset in the community, consisting of 3.6 million images distributed into seven folds. , According to previous research \cite{jiang2021regressive,kim2022unified}, S1, S5, S6, S7, and S8 are designated as the training set, whereas S9 and S11 serve as the testing set. The other target dataset is \textbf{Leeds Sports Pose} \cite{johnson2010clustered} (LSP), which is a real-world dataset with 2,000 images, and we use all of them for adaptation. We focus on two domain adaptation tasks: SURREAL $\rightarrow$ Human3.6M and SURREAL $\rightarrow$ LSP for human pose tasks.

\textbf{Rendered Hand Pose Dataset} \cite{zimmermann2017learning} (RHD) is the source hand dataset with 43,986 synthetic hand images,  of which 41,258 are allocated for training, while the remaining 2,728 images are utilized for validation. \textbf{Hand-3D-Studio} \cite{zhao2020h3d} (H3D) is one of the two target hand datasets, including 22,000 real-world frames. For training, we employed 18,800 frames and reserved the rest for testing, following the protocol outlined in previous research papers \cite{jiang2021regressive, kim2022unified}. The other target hand dataset, FreiHand \cite{zimmermann2019freihand}, provides 130k real-world images, and we employed all of them for our adaptation task. Our focus is on two domain adaptation tasks: RHD $\rightarrow$ H3D and RHD $\rightarrow$ FreiHand.

\textbf{\textcolor{blue}{Implementation Details.}} We adopt Simple Baseline \cite{xiao2018simple} with ResNet-101 \cite{he2016deep} as the HPE backbone, following previous works \cite{jiang2021regressive,kim2022unified}. In the \emph{pretrain} process, we conduct 40 epochs, each with 500 iterations, using the Adam \cite{kingma2014adam} optimizer with an initial learning rate of 1e-4. The learning rate decreases to 1e-5 at 25 epochs. For the \emph{adaptation} process, we execute 40 epochs with 30,000 iterations. In step A, we choose an initial learning rate of 1e-4 for $F^{sr}$ and 1e-5 for $G^{in}$. Step B sets the initial learning rate to 1e-4 for $G^{tg}$ and 1e-3 for $F^{tg}$. We use the same annealing strategy for the learning rate scheduler as in \cite{jiang2021regressive}. As for hyperparameters, we select $\alpha=0.7$, $\beta=0.5$, and $\gamma=0.85$. Moreover, the intermediate model is used for the final inference.

%As for hyperparameters, we set $\alpha_{1} = \alpha_{2} = 0.5$, and select $\beta$ as $0.2$. Besides, $\gamma_{1}$ and $\gamma_{2}$ are set to be $0.4$ and $0.7$. To represent the maximization process of Stage B better, we use the same strategy as \cite{jiang2021regressive} to minimize negative heatmaps built from spatial probability. All experiments are conducted on Nvidia RTX A5000 GPUs. 

\subsection{Main Results}

\textbf{\textcolor{blue}{Baselines.}}  Four general domain adaptive pose estimation methods \textbf{CC-SSL} \cite{mu2020learning}, \textbf{MDAM} \cite{li2021synthetic}, \textbf{RegDA} \cite{jiang2021regressive} and \textbf{UniFrame} \cite{kim2022unified} with SOTA performances are chosen as baselines for comparison. We list these baselines that need to access source data to demonstrate that our proposed source-free method's performance remains competitive when compared to general domain adaptive pose estimation methods. We build source-free baselines by either adapting domain-adaptive HPE methods or source-free DA methods for classification methods. We first introduce two domain adaptive HPE methods. \textbf{RegDA-SF} is based on \cite{jiang2021regressive}, which originally contains three steps and the first step uses source data. We replace the first step with an exponential moving averaging strategy to obtain the new method. \textbf{Uniframe-SF} is on the foundation of \cite{kim2022unified}. What we do is remove the style transfer modules between source and target so that it becomes a source-free method. The rest are adapted from source-free DA methods for classification methods. \textbf{SHOT} \cite{liang2020we} is a classic source-free domain adaptation method for image classification, and we adapt its mutual information maximization module for HPE by treating the normalized heatmaps as output probabilities in classification. \textbf{MMT} is based on \cite{ge2020mmt}, which is a source-free adaptation method that applies two models mutually learning from each other. Here we replace its soft classification loss with the soft consistency loss. \textbf{SHOT++} \cite{liang2022shot++} is the advanced version of \cite{liang2020we}, which uses a special unsupervised representation learning technique predicting image rotations. We also add this module to our adopted baseline. %Additionally, since the source-free domain adaptive HPE setting is a fairly new field, we come up with several baselines. \textbf{RegDA-SF} is based on \cite{jiang2021regressive}, which originally contains three steps and the first step uses source data. We replace the first step with an exponential moving averaging strategy to obtain the new method. \textbf{SHOT} \cite{liang2020we} is a classic source-free domain adaptation method for image classification, and we adapt its mutual information maximization module for HPE by treating the normalized heatmaps as output probabilities in classification. \textbf{MMT} is based on \cite{ge2020mmt}, which is a source-free adaptation method that applies two models mutually learning from each other. Here we replace its soft classification loss with the soft consistency loss. \textbf{Uniframe-SF} is on the foundation of \cite{kim2022unified}. What we do is remove the style transfer modules between source and target. Actually, it turns out to be the application of Mean Teacher \cite{tarvainen2017mean} in HPE. \textbf{SHOT++} \cite{liang2022shot++} is the advanced version of \cite{liang2020we}, which uses a special unsupervised representation learning technique predicting image rotations. We also add this module to our adopted baseline. Besides, we add \textbf{Source-only} as another baseline for all four tasks to evaluate the effectiveness of adaptation.   

\textbf{\textcolor{blue}{Metrics.}} To evaluate the accuracy of the proposed approach, we adopt the Percentage of Correct Keypoint (PCK) metric, as used in previous work. We set the ratio of correct predictions to be $5\%$, and report PCK@0.05 in Tables \ref{tab:rhd2h3d}-\ref{tab:surreal2lsp}. In addition to overall keypoint accuracy, we divide joints into several part segments and measure their performance using specific metrics. For the 21-keypoint hand skeleton, we use the Metacarpophalangeal (MCP), Proximal Interphalangeal (PIP), Distal Interphalangeal (DIP), and Fingertip (Fin) metrics. For the 18-keypoint human skeleton, we select the Shoulder (Sld), Elbow (Elb), Wrist, Hip, Knee, and Ankle metrics. These segment-specific metrics enable a more detailed evaluation of the models' performances.
%\vspace{-5pt}
\begin{table}[!ht]
    \scriptsize
    \centering
    \caption{PCK@0.05 on RHD $\rightarrow$ H3D Task}%\vspace{-1mm}
    \resizebox{0.97\linewidth}{!}{%
    \begin{tabular}{rcccccccc}
          \toprule
          Method & SF &  MCP &  PIP  &  DIP & Fin & All \\
         \hline
         {Source-only} & - & 67.4 & 64.2 & 63.3 & 54.8 & 61.8\\
         %{Oracle} & 97.7 & 97.2 & 95.7 & 92.5 & 95.8\\
         \hline
         CC-SSL \cite{mu2020learning} (CVPR'20) & $\times$ & 81.5 & 79.9 & 74.4 & 64.0 & 75.1 \\
         MDAM \cite{li2021synthetic} (CVPR'21) & $\times$ & 82.3 & 79.6 & 72.3 & 61.5 & 74.1\\
         RegDA \cite{jiang2021regressive} (CVPR'21) & $\times$ & 79.6 & 74.4 & 71.2 & 62.9 & 72.5\\
         UniFrame \cite{kim2022unified} (ECCV'22) & $\times$ & 86.7 & 84.6 & 78.9 & 68.1 & 79.6\\
         \hline 
         RegDA-SF \cite{jiang2021regressive} (CVPR'21) & \checkmark & 71.2 & 66.8 & 66.2 & 58.5 & 66.9 \\  UniFrame-SF \cite{kim2022unified} (ECCV'22) & \checkmark & 84.8 & 84.2 & 77.0 & 68.0 & 77.9\\
         SHOT \cite{liang2020we} (ICML'20) & \checkmark & 77.5 & 69.8 & 70.6 & 64.2 & 73.7 \\
         MMT \cite{ge2020mmt} (ICLR'20) & \checkmark & 78.7 & 74.1 & 72.5 & 65.3 & 75.3\\
         SHOT++ \cite{liang2022shot++} (TPAMI'22) & \checkmark & 85.3 & 85.1 & 78.2 & 67.6 & 78.9\\
         {Ours} & \checkmark &  \textbf{88.4} & \textbf{89.2} & \textbf{80.9} & \textbf{71.4} & \textbf{82.2}\\    
         \bottomrule
    \end{tabular}%
    }
    %\vspace{-15pt}
\label{tab:rhd2h3d}
\end{table}

%\vspace{-5pt}
\begin{table}[!ht]
    \scriptsize
    \centering
    \caption{PCK@0.05 on RHD $\rightarrow$ FreiHand Task}%\vspace{-1mm}
    \resizebox{0.97\linewidth}{!}{%
    \begin{tabular}{rcccccccc}
          \toprule
          Method & SF & MCP &  PIP  &  DIP & Fin & All \\
         \hline
         {Source-only} & - & 35.2 & 50.1 & 54.8 & 50.7 & 46.8\\
         \hline
         CC-SSL \cite{mu2020learning} (CVPR'20) & $\times$ & 37.4 & 48.2 & 50.1 & 46.5 & 43.8\\
         MDAM \cite{li2021synthetic} (CVPR'21) & $\times$ & 32.3 & 48.1 & 51.7 & 47.3 & 45.1\\
         RegDA \cite{jiang2021regressive} (CVPR'21) & $\times$  & 40.9 & 55.0 & 58.2 & 53.1 & 51.1\\
         UniFrame \cite{kim2022unified} (ECCV'22) & $\times$  & 43.5 & 64.0 & 67.4 & 62.4 & 58.5\\
         \hline
         RegDA-SF \cite{jiang2021regressive} (CVPR'21) & \checkmark &  38.6 & 52.9 & 57.6 & 54.3 & 49.5\\  UniFrame-SF \cite{kim2022unified} (ECCV'22) & \checkmark & 40.6 & 62.5 & 61.0 & 60.2 & 55.7\\     
         SHOT \cite{liang2020we} (ICML'20) & \checkmark & 40.4 & 61.3 & 60.5 & 58.1 & 53.3\\
         MMT \cite{ge2020mmt} (ICLR'20) & \checkmark & 39.6 & 60.4 & 60.0 & 57.8 & 52.6\\
         
         SHOT++ \cite{liang2022shot++} (TPAMI'22) & \checkmark & 41.0 & 62.8 & 62.7 & 59.3 & 55.8\\
         {Ours} & \checkmark & \textbf{43.7} & \textbf{65.9} & \textbf{66.6} & \textbf{63.1} & \textbf{58.8}\\    
         \bottomrule
    \end{tabular}%
    }
    %\vspace{-10pt}
\label{tab:rhd2fre}
\end{table}

\begin{table}[!ht]
    \scriptsize
    \centering
    \caption{PCK@0.05 on SURREAL $\rightarrow$ Human3.6M Task}%\vspace{-1mm}
    \resizebox{0.97\linewidth}{!}{%
    \begin{tabular}{rcccccccc}
          \toprule
          Method & SF &Sld & Elb & Wrist & Hip & Knee & Ankle & All \\
         \hline
         {Source-only} & - & 69.4 & 75.4 & 66.4 & 37.9 & 77.3 & 77.7 & 67.3\\
         %{Oracle} & 95.3 & 91.8 & 86.9 & 95.6 & 94.1 & 93.6 & 92.9\\
         \hline
         CC-SSL \cite{mu2020learning} (CVPR'20) & $\times$ & 44.3 & 68.5 & 55.2 & 22.2 & 62.3 & 57.8 & 51.7 \\
         MDAM \cite{li2021synthetic} (CVPR'21) & $\times$ & 51.7 & 83.1 & 68.9 & 17.7 & 79.4 & 76.6 & 62.9\\
         RegDA \cite{jiang2021regressive} (CVPR'21) & $\times$ & 73.3 & 86.4 & 72.8 & 54.8 & 82.0 & 84.4 & 75.6\\
         UniFrame \cite{kim2022unified} (ECCV'22) & $\times$ & 78.1 & 89.6 & 81.1 & 52.6 & 85.3 & 87.1 & 79.0\\
         \hline
         RegDA-SF \cite{jiang2021regressive} (CVPR'21) & \checkmark &  70.6 & 82.0 & 69.8 & 43.3 & 79.1 & 79.4 & 71.5\\
         UniFrame-SF \cite{kim2022unified} (ECCV'22) & \checkmark & 74.3 & 84.8 & 72.5 & 45.9 & 81.5 & 85.2 & 75.1\\         
         SHOT \cite{liang2020we} (ICML'20) & \checkmark &  73.0 & 83.2 & 71.9 & 46.7 & 79.8 & 82.4 & 73.3\\
         MMT \cite{ge2020mmt} (ICLR'20) & \checkmark & 73.2 & 83.5 & 72.4 & 45.1 & 80.8 & 83.9 & 73.9\\
         SHOT++ \cite{liang2022shot++} (TPAMI'22) & \checkmark & 75.0 & 85.2 & 76.7 & 45.3 & 82.8 & 85.0 & 75.3\\
         {Ours} & \checkmark &  \textbf{77.9} & \textbf{88.8} & \textbf{80.4} & \textbf{52.3} & \textbf{84.2} & \textbf{86.9} & \textbf{78.7}\\       
         \bottomrule
    \end{tabular}%
    }
    %\vspace{-10pt}
\label{tab:surreal2h36m}
\end{table}

\begin{table}[!ht]
    \scriptsize
    \centering
    \caption{PCK@0.05 on SURREAL $\rightarrow$ LSP  Task}      %\vspace{-1mm}
    \resizebox{0.97\linewidth}{!}{%
    \begin{tabular}{rcccccccc}
          \toprule
          Method & SF & Sld & Elb & Wrist & Hip & Knee & Ankle & All \\
         \hline
         {Source-only} & - & 51.5 & 65.0 & 62.9 & 68.0 & 68.7 & 67.4 & 63.9 \\
         \hline
         CC-SSL \cite{mu2020learning} (CVPR'20) & $\times$ & 36.8 & 66.3 & 63.9 & 59.6 & 67.3 & 70.4 & 60.7 \\
         MDAM \cite{li2021synthetic} (CVPR'21) & $\times$ & 61.4 & 77.7 & 75.5 & 65.8 & 76.7 & 78.3 & 69.2 \\
         RegDA \cite{jiang2021regressive} (CVPR'21) & $\times$ & 62.7 & 76.7 & 71.1 & 81.0 & 80.3 & 75.3 & 74.6\\
         UniFrame \cite{kim2022unified} (ECCV'22) & $\times$ & 69.2 & 84.9 & 83.3 & 85.5 & 84.7 & 84.3 & 82.0\\
         \hline 
         RegDA-SF \cite{jiang2021regressive} (CVPR'21) & \checkmark & 54.8 & 70.5 & 67.6 & 65.4 & 73.2 & 70.0 & 66.5\\
         UniFrame-SF \cite{kim2022unified} (ECCV'22) & \checkmark & 68.4 & 80.5 & 79.1 & 82.7 & 80.8 & 81.0 & 78.8\\
         SHOT \cite{liang2020we} (ICML'20) & \checkmark & 63.5 & 72.7 & 66.5 & 78.4 & 79.7 & 73.2 & 72.4\\
         MMT \cite{ge2020mmt} (ICLR'20) & \checkmark & 60.9 & 70.9 & 70.3 & 81.1 & 79.3 & 72.8 & 71.5\\
         
         SHOT++ \cite{liang2022shot++} (TPAMI'22) & \checkmark & 69.5 & 81.7 & 80.9 & 84.0 & 82.3 & 79.7 & 79.9\\
         {Ours} & \checkmark &  \textbf{70.7} & \textbf{85.4} & \textbf{83.8} & \textbf{86.6} & \textbf{85.2} & \textbf{85.0} & \textbf{83.2}\\   
         \bottomrule
    \end{tabular}%
    }
%\vspace{-15pt}
\label{tab:surreal2lsp}
\end{table}

\begin{figure*}[ht]
  \centering
   \includegraphics[width=1.00\linewidth]{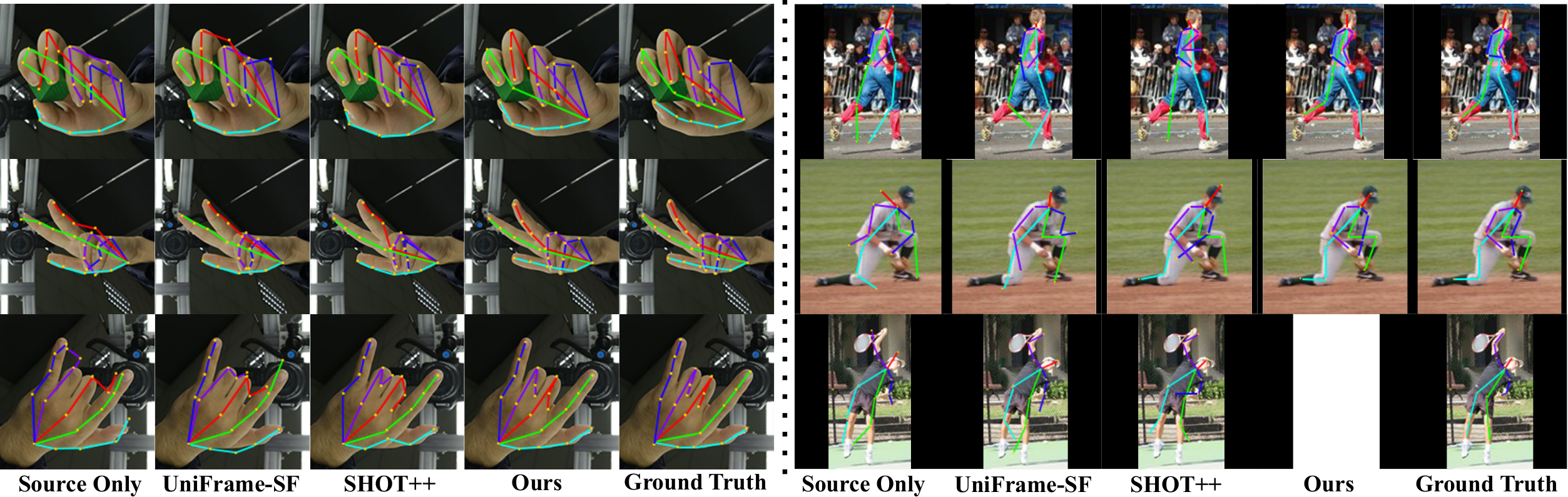}
    %\vspace{-15pt}
   \caption{ Qualitative results on H3D dataset (left) and LSP dataset (right).
   }%\vspace{-10pt}
   \label{fig:vis}
\end{figure*}

\textbf{\textcolor{blue}{Quantitative Results.}} We present the results of our proposed method on four different tasks: hand tasks RHD$\rightarrow$H3D and RHD$\rightarrow$FreiHand, and human tasks SURREAL$\rightarrow$Human3.6M and SURREAL$\rightarrow$LSP. Tables \ref{tab:rhd2h3d} and \ref{tab:rhd2fre} show the results for the hand tasks, where our method achieves state-of-the-art performance, surpassing the second best baseline SHOT++ with a significant lead of $3.3\%$ and $3.0\%$, respectively. Moreover, our method outperforms the SHOT++ model by $4.1\%$ and $3.9\%$ on the PIP metric, respectively. For the human tasks, Tables \ref{tab:surreal2h36m} and \ref{tab:surreal2lsp} display the results. Our model achieves a considerable margin of $3.4\%$ and $3.3\%$ over the second best baseline SHOT++ for SURREAL$\rightarrow$Human3.6M and SURREAL$\rightarrow$LSP, respectively. Specifically, our method outperforms SHOT++ by $7.0\%$ on the Wrist for the adaptation on Human3.6M and $5.3\%$ on the Ankle for the adaptation on LSP. Furthermore, our approach remains competitive when compared with general domain adaptive HPE methods that need to access source data, as it surpasses UniFrame by $2.6\%$ on RHD$\rightarrow$H3D, $0.3\%$ on RHD$\rightarrow$FreiHand, and $1.2\%$ on SURREAL$\rightarrow$LSP.%Only on the adaptation on Human3.6M, our method falls slightly behind Uniframe by $0.3\%$.

\iffalse
\begin{figure}[t]
  \centering
   \includegraphics[width=1.00\linewidth]{img/h3d001.pdf}

   \caption{ Qualitative results on H3D dataset
   } \vspace{-10pt}
   \label{fig:h3d}
\end{figure}

\begin{figure}[t]
  \centering
   \includegraphics[width=1.00\linewidth]{img/h36m001.pdf}

   \caption{ Qualitative results on Human3.6M dataset
   }\vspace{-15pt}
   \label{fig:h36m}
\end{figure}
\fi

\textbf{\textcolor{blue}{Qualitative Results. }} Figs. \ref{fig:vis} shows qualitative results of H3D on the left side and LSP on the right side. We use Source only, \textbf{UniFrame-SF} \cite{kim2022unified}, \textbf{SHOT++} \cite{liang2022shot++}, \textbf{Ours}, and \textbf{Ground Truth} for qualitative comparison%, and more results on other tasks are in \textcolor{blue}{supplementary}
. It is evident that our method outperforms other baselines significantly.

\subsection{Ablation Study on Framework}
\label{sec:frame}

Our method contains two modules in Step A \& Step B as the source-protect module (\textbf{SP}) and the target-relevant module (\textbf{TR}) separately, and here we focus on their functions. Moreover, we use \textbf{MMT} \cite{ge2020mmt} as the baseline, which is referred in Table \ref{tab:rhd2h3d}-\ref{tab:surreal2lsp} . Table \ref{tab:rh-ab} and \ref{tab:sh-ab} show the ablation study of frameworks on RHD$\rightarrow$H3D and SURREAL$\rightarrow$Human3.6M with different combinations of these modules. %More results on other tasks are in \textcolor{blue}{supplementary}. %In this part, we focus on the structures of preserving source information and provide several baselines. 
%\textbf{PM} means using the model adapted in the previous iteration for source knowledge distilling, which is applied in SHOT and SHOT++. \textbf{MT} is Mean Teacher and applies EMA to protect source pretrained model as Uniframe-SF does. 
%\textbf{MMT} is Mutual Mean Teaching which is identical to the baseline we used in Table \ref{tab:rhd2h3d}-\ref{tab:surreal2lsp} and also contains EMA. Our design for maintaining source knowledge is \textbf{SP} (the source-protect module), and for the fairness of comparisons, we add \textbf{TR} (the target-relevant module) to all these frameworks.
%\textbf{IDF} means intermediate domain framework, which is the framework proposed in Fig. \ref{fig:framework}. \textbf{DL} corresponds to the new discrepancy loss shown in Section \ref{sec:dl}, which is in close cooperation with IDF. \textbf{RW} is the reweighting loss proposed in Section \ref{sec:rw}. 

%From these tables, we observe that with only IDF and no additional modules, the model cannot show superiority. But when IDF meets DL, they boost performance greatly. That is because the intermediate characteristic is maintained by both IDF and DL, and removing any of them will destroy the construction of intermediate representations. We observe that PCK@0.05 increases $7.9\%$ for RHD$\rightarrow$H3D, and $4.1\%$ for SURREAL$\rightarrow$Human3.6M with the assistance of DL. The improvements from RW are not as great as DL but are still obvious. With the help of RW, the performance gains $1.8\%$ on both RHD$\rightarrow$H3D and SURREAL$\rightarrow$Human3.6M. 

%\ke{baseline without each component}
\begin{table}[!ht]
    \scriptsize
    \centering
    \caption{Ablation of Frameworks on RHD $\rightarrow$ H3D}
    \resizebox{1.0\linewidth}{!}{%
    \begin{tabular}{ccccccccc}
          \toprule
          Method &  MCP &  PIP  &  DIP & Fin & All \\
         \hline
         %{PM+PS} &  78.9 & 70.4 & 71.9 & 65.2 & 74.8\\
         {MMT \cite{ge2020mmt}} & 78.7 & 74.1 & 72.5 & 65.3 & 75.3\\
         {MMT \cite{ge2020mmt} +TR} & 81.3 & 77.5 & 76.0 & 67.7 & 79.1\\
         %{MT+PS} & 86.1 & 84.8 & 77.9 & 69.2 & 80.1\\ 
         {SP+TR (Ours)} & 88.4 & 89.2 & 80.9 & 71.4 & 82.2\\
         \bottomrule
    \end{tabular}%
    }
    %\vspace{-5pt}
\label{tab:rh-ab}
\end{table}

%\vspace{-8pt}
\begin{table}[!ht]
    \scriptsize
    \centering
    \caption{Ablation of Frameworks on SURREAL $\rightarrow$ Human3.6M}
    \resizebox{1.0\linewidth}{!}{%
    \begin{tabular}{ccccccccc}
          \toprule
          Method &  Sld & Elb & Wrist & Hip & Knee & Ankle & All \\
         \hline
         %{PM+PS} & 74.5 & 84.0 & 72.7 & 47.6 & 81.5 & 83.2 & 74.4\\
         {MMT \cite{ge2020mmt}} & 73.2 & 83.5 & 72.4 & 45.1 & 80.8 & 83.9 & 73.9\\
         {MMT \cite{ge2020mmt} +TR} & 74.4 & 86.5 & 75.7 & 48.9 & 81.3 & 85.1 & 76.2 \\
         %{MT+PS} & 75.0 & 85.6 & 73.7 & 48.8 & 82.5 & 85.4 & 76.4\\ 
         {SP+TR (Ours)} & 77.9 & 88.8 & 80.4 & 52.3 & 84.2 & 86.9 & 78.7\\   
         \bottomrule
    \end{tabular}%
    }
    %\vspace{-10pt}
\label{tab:sh-ab}
\end{table}

The results clearly demonstrate that both SP and TR contribute to improving the model's performance. Specifically, TR enhances the model's accuracy by $3.8\%$ on RHD $\rightarrow$ H3D and $2.3\%$ on SURREAL $\rightarrow$ Human3.6M, while SP leads to an improvement of $4.1\%$ on RHD $\rightarrow$ H3D and $2.5\%$ on SURREAL $\rightarrow$ Human3.6M compared to MMT. Notably, the two proposed modules provide similar levels of improvement.

%Based on these results, it is evident that both SP and TR play a role in improving the model's performance. We can see that TR improves $3.8\%$ on RHD $\rightarrow$ H3D and $2.3\%$ on SURREAL $\rightarrow$ Human3.6M when working with MMT. As for SP, it improves $4.1\%$ on RHD $\rightarrow$ H3D and $2.5\%$ on SURREAL $\rightarrow$ Human3.6M when compared with MMT. Moreover, we observe that the significance of these two proposed modules are almost the same.
%Based on these results, it is evident that our source-protect module surpasses all other source-preserving structures with a margin of $2.1\%$ on RHD$\rightarrow$H3D and $2.3\%$ on SURREAL$\rightarrow$Human3.6M, thereby demonstrating the superiority of our proposed module.
%\vspace{-8pt}

%\vspace{-9pt}
\subsection{Ablation Study on Proposed Losses}

%\ZD{Too many tables for Ablation study. Move some to SM.}
%We further conduct a detailed ablation study on the three losses proposed $\mathcal{L}_{res}$, $\mathcal{L}_{cst}$ and $\mathcal{L}_{im}$ using two tasks RHD$\rightarrow$H3D and SURREAL$\rightarrow$Human3.6M, as shown in Table \ref{tab:rh-ab-ls} and Table \ref{tab:sh-ab-ls}. Since we have already discussed the functions of the source-protect module and target-relevant module, and show their superiority in Sec. \ref{sec:frame},  $\mathcal{L}_{ft}$ from the source-protect module and $\mathcal{L}_{con}$ from the target-relevant module are kept in the \textbf{Baseline} here.
We performed a detailed ablation study on the three proposed losses, namely $\mathcal{L}_{res}$, $\mathcal{L}_{cst}$, and $\mathcal{L}_{im}$, using the RHD$\rightarrow$H3D and SURREAL$\rightarrow$Human3.6M tasks. Tables \ref{tab:rh-ab-ls} and \ref{tab:sh-ab-ls} present the results. As we have previously discussed the functions and advantages of the source-protect and target-relevant modules in Sec. \ref{sec:frame}, we retained $\mathcal{L}_{ft}$ from the source-protect module and $\mathcal{L}_{con}$ from the target-relevant module in the \textbf{Baseline}.
%According to the three relations described in Section \ref{sec:dl}, four combinations $r_1$, $r_1$ \& $r_2$,  $r_1$ \& $r_3$ and $r_1$ \& $r_2$ \& $r_3$ are implemented and the results are listed in Table \ref{tab:rh-ab-dl} and \ref{tab:sh-ab-dl}. We observe that the removal of either $r_2$ or $r_3$ will degrade the model's performance. Simply removing $r_2$ leads to a decrease of $2.9\%$ in RHD$\rightarrow$H3D and $2.3\%$ in SURREAL$\rightarrow$Human3.6M. As for $r_3$, eliminating it causes a drop of $1.3\%$ in RHD$\rightarrow$H3D and $1.1\%$ in SURREAL$\rightarrow$Human3.6M. Besides, it is noticed that $r_2$ plays a more important role than $r_3$.  

%\vspace{-7pt}
\begin{table}[!ht]
    \scriptsize
    \centering
    \caption{Ablation of Losses on RHD $\rightarrow$ H3D}
    \resizebox{1.0\linewidth}{!}{%
    \begin{tabular}{ccccccccc}
          \toprule
          Method &  MCP &  PIP  &  DIP & Fin & All \\
         \hline
         {Baseline} & 85.3 & 86.8 & 78.5 & 68.1 & 79.4\\
         {$\mathcal{L}_{res}$} & 85.5 & 87.9 & 79.3 & 70.7 & 80.1\\
         {$\mathcal{L}_{cst}$} & 85.5 & 89.0 & 78.4 & 70.1 & 80.9\\
         {$\mathcal{L}_{im}$} & 85.8 & 87.2 & 78.7 & 69.8 & 80.3\\
         {$\mathcal{L}_{cst} \& \mathcal{L}_{im}$} & 86.9 & 88.8 & 80.1 & 70.3 & 81.4\\
         {$\mathcal{L}_{res} \& \mathcal{L}_{cst} \& \mathcal{L}_{im}$} & 88.4 & 89.2 & 80.9 & 71.4 & 82.2\\
         \bottomrule
    \end{tabular}%
    }
    %\vspace{-10pt}
\label{tab:rh-ab-ls}
\end{table}

%\vspace{-5pt}
\begin{table}[!ht]
    \scriptsize
    \centering
    \caption{Ablation of Losses on SURREAL $\rightarrow$ Human3.6M}
    \resizebox{1.0\linewidth}{!}{%
    \begin{tabular}{ccccccccc}
          \toprule
          Method &  Sld & Elb & Wrist & Hip & Knee & Ankle & All \\
         \hline
         {Baseline} & 75.4 & 86.1 & 76.8 & 46.5 & 83.0 & 85.0 & 75.8\\
         {$\mathcal{L}_{res}$} & 76.0 & 86.9 & 77.5 & 47.8 & 83.3 & 85.2 & 76.6\\
         {$\mathcal{L}_{cst}$} & 76.6 & 87.4 & 78.8 & 50.6 & 83.5 & 86.0 & 77.2 \\
         {$\mathcal{L}_{im}$} & 76.2 & 87.1 & 78.3 & 49.7 & 83.2 & 85.7 & 76.9 \\
         {$\mathcal{L}_{cst} \& \mathcal{L}_{im}$} & 77.3 & 87.7 & 79.2 & 51.2 & 83.6 & 86.4 & 78.0\\
         {$\mathcal{L}_{res} \& \mathcal{L}_{cst} \& \mathcal{L}_{im}$} & 77.9 & 88.8 & 80.4 & 52.3 & 84.2 & 86.9 & 78.7\\  
         \bottomrule
    \end{tabular}%
    }
    %\vspace{-10pt}
\label{tab:sh-ab-ls}
\end{table}

We observe that each loss is able to boost the model's performance. Simply applying $\mathcal{L}_{res}$ leads to a increase of $0.7\%$ in RHD$\rightarrow$H3D and $0.8\%$ in SURREAL$\rightarrow$Human3.6M. $\mathcal{L}_{cst}$ causes an improvement of $1.5\%$ in RHD$\rightarrow$H3D and $1.4\%$ in SURREAL$\rightarrow$Human3.6M. As for $\mathcal{L}_{im}$, adding it causes a rise of $0.9\%$ in RHD$\rightarrow$H3D and $1.1\%$ in SURREAL$\rightarrow$Human3.6M. Besides, it is noticed that $\mathcal{L}_{cst}$ plays a more important role than $\mathcal{L}_{res}$ or $\mathcal{L}_{im}$.

\iffalse
\begin{figure}[!ht]
%\vspace{-5pt}
  \centering
  \includegraphics[width=1\linewidth]{img/params003.pdf}%\vspace{-1mm}
  \caption{Parameter Analysis on RHD$\rightarrow$H3D (best viewed in color). \textbf{a}: Analysis on $\alpha$. \textbf{b}: Analysis on $\beta$. \textbf{c}: Analysis on $\gamma$. 
   }%\vspace{-10pt}
   \label{fig:param}
\end{figure}
\fi

\subsection{Ablation of Sparsity Reduction}
\label{sec:ab-pro}

In this section, we investigate the effectiveness of our proposed sparsity reduction strategy. Specifically, we compare heatmap-based methods and vector-based methods for two loss functions: contrastive learning ($\mathcal{L}_{cst}$) and information maximization ($\mathcal{L}_{im}$). We denote heatmap-based contrastive learning and information maximization as \textbf{HBCL} and \textbf{HBIM}, respectively, and vector-based contrastive learning and information maximization as \textbf{VBCL} and \textbf{VBIM}, respectively. Two tasks are selected and the results of these comparisons are presented in Tab. \ref{tab:rh-ab-sr} and Tab. \ref{tab:sh-ab-sr}.

\begin{table}[!ht]
    \scriptsize
    \centering
    \caption{Ablation of Sparsity Reduction on RHD $\rightarrow$ H3D}
    \resizebox{1.0\linewidth}{!}{%
    \begin{tabular}{ccccccccc}
          \toprule
          Method &  MCP &  PIP  &  DIP & Fin & All \\
         \hline
         %{PM+PS} &  78.9 & 70.4 & 71.9 & 65.2 & 74.8\\
         {MMT \cite{ge2020mmt}}  & 78.7 & 74.1 & 72.5 & 65.3 & 75.3\\
         \hline
         {MMT \cite{ge2020mmt} + HBCL} & 77.4 & 73.0 & 70.7 & 65.5 & 74.2\\
         {MMT \cite{ge2020mmt} + VBCL} & \textbf{81.7} & \textbf{76.6} & \textbf{74.4} & \textbf{67.9} & \textbf{78.7} \\
         \hline
         {MMT \cite{ge2020mmt} + HBIM} & 79.3 & 74.7 & 72.9 & 65.8 & 75.9\\
         {MMT \cite{ge2020mmt} + VBIM} & \textbf{80.3} & \textbf{76.1} & \textbf{73.6} & \textbf{66.5} & \textbf{77.5}\\
         \bottomrule
    \end{tabular}%
    }
    %\vspace{-5pt}
\label{tab:rh-ab-sr}
\end{table}

\begin{table}[!ht]
    \scriptsize
    \centering
    \caption{Ablation of Sparsity Reduction on SURREAL $\rightarrow$ Human3.6M}
    \resizebox{1.0\linewidth}{!}{%
    \begin{tabular}{ccccccccc}
          \toprule
          Method &  Sld & Elb & Wrist & Hip & Knee & Ankle & All \\
         \hline
         %{PM+PS} & 74.5 & 84.0 & 72.7 & 47.6 & 81.5 & 83.2 & 74.4\\
         {MMT \cite{ge2020mmt}} & 73.2 & 83.5 & 72.4 & 45.1 & 80.8 & 83.9 & 73.9\\
         \hline
         {MMT \cite{ge2020mmt} + HBCL} & 71.7 & 82.0 & 71.8 & 44.6 & 80.2 & 82.4 & 73.0\\
         {MMT \cite{ge2020mmt} + VBCL} & \textbf{75.6} & \textbf{87.0} & \textbf{75.1} & \textbf{49.2} & \textbf{83.3} & \textbf{86.3} & \textbf{76.7}\\
         \hline
         {MMT \cite{ge2020mmt} + HBIM} & 73.9 & 84.6 & 72.7 & 46.0 & 81.1 & 83.0 & 74.3\\
         {MMT \cite{ge2020mmt} + VBIM} & \textbf{75.3} & \textbf{86.4} & \textbf{74.3} & \textbf{48.6} & \textbf{82.4} & \textbf{85.7} & \textbf{76.0}\\  
         \bottomrule
    \end{tabular}%
    }
    %\vspace{-10pt}
\label{tab:sh-ab-sr}
\end{table}

%Based on the results, we can observe that HBCL does not improve the model's performance, whereas VBCL provides significant assistance. For example, with the combination of HBCL, MMT+HBCL decreases $1.1\%$ when compared with MMT on the task RHD $\rightarrow$ H3D, while MMT+VBCL increases $3.4\%$ when compare with MMT. Additionally, although HBIM can enhance the model's performance, it cannot surpass VBIM. For example, with the combination of HBCL, MMT+HBIM increases $0.6\%$ when compared with MMT on the task RHD $\rightarrow$ H3D, while MMT+VBIM increases $2.2\%$ when compare with MMT. To summarize, it is evident that utilizing the vector-based spatial probability space is essential for achieving optimal results.

Based on the results, it is evident that HBCL does not improve the model's performance significantly, whereas VBCL provides significant assistance. For instance, MMT+HBCL shows a decrease of $1.1\%$ on the RHD $\rightarrow$ H3D task, while MMT+VBCL exhibits an increase of $3.4\%$. Additionally, while HBIM can enhance the model's performance, it falls short of surpassing VBIM. For example, MMT+HBIM shows an increase of $0.6\%$ on the RHD $\rightarrow$ H3D task, whereas MMT+VBIM achieves an increase of $2.2\%$. In summary, it becomes evident that utilizing the vector-based spatial probability space is crucial for achieving optimal results.

\subsection{Parameter Analysis}

\begin{figure}[!ht]
%\vspace{-5pt}
  \centering
  \includegraphics[width=1\linewidth]{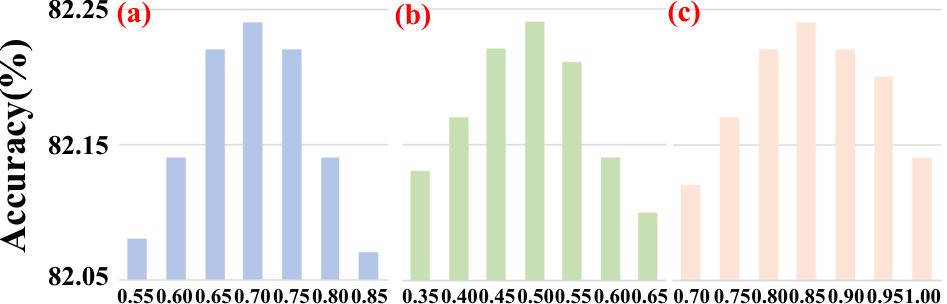}%\vspace{-1mm}
  \caption{Parameter Analysis on RHD$\rightarrow$H3D (best viewed in color). \textbf{a}: Analysis on $\alpha$. \textbf{b}: Analysis on $\beta$. \textbf{c}: Analysis on $\gamma$. 
   }%\vspace{-10pt}
   \label{fig:param}
\end{figure}

We use RHD$\rightarrow$H3D to illustrate the sensitivity of $\alpha$ in Equation \ref{eq:in} and $\beta$ and $\gamma$ in Equation \ref{eq:tgt}. The results are shown in Fig. \ref{fig:param}. From it, we observe that $\alpha=0.7$, $\beta=0.5$, and $\gamma=0.85$ is the best choice. Moreover, it is noticed that the decreases from different parameters are limited to $0.2\%$ for $\alpha$, $\beta$ and $\gamma$. Therefore, our method shows stable performance over hyperparameters.

\section{Qualitative Results on the Unseen Dataset COCO}
\label{sec:coco}

\begin{figure}[!ht]
  \centering
   \includegraphics[width=1.0\linewidth]{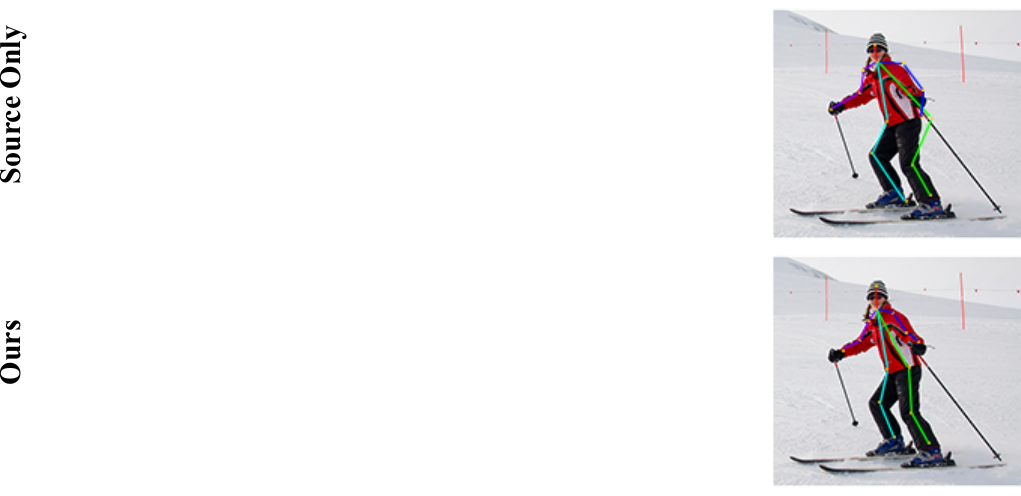}
   \caption{ Qualitative results on the COCO dataset. Here we compare our method trained from SURREAL $\rightarrow$ LSP with the source-only model.
   }
   \label{fig:coco}
\end{figure}

%This section presents qualitative results on the COCO dataset \cite{lin2014microsoft}, which was not used in our previous human dataset studies. That is because our 14-keypoint poses are annotated in a different way with the 17-keypoint poses in COCO, and that's the reason why we cannot provide quantitative results on COCO with its own metrics or APIs.  Although we cannot provide quantitative generalization results, we believe that our qualitative results are still significant. Our analysis employs the model trained on the SURREAL $\rightarrow$ LSP task in combination with the pre-trained SURREAL model. The results, presented in Fig. \ref{fig:coco}, illustrate the model's ability to generalize to unseen datasets, suggesting that the adaptation process enhances the model's generalization ability.

In this section, we showcase qualitative results on the COCO dataset \cite{lin2014microsoft}, which was not utilized in our previous human dataset studies. It's important to note that our 14-keypoint poses are annotated differently from the 17-keypoint poses in COCO, which hinders providing quantitative results on COCO using its own metrics or APIs. Despite this limitation, we believe that our qualitative findings hold significant value. For our analysis, we leverage the model trained on the SURREAL $\rightarrow$ LSP task in conjunction with the pre-trained SURREAL model. The results, presented in Fig. \ref{fig:coco}, demonstrate the model's capacity to generalize effectively to unseen datasets, implying that the adaptation process boosts the model's generalization ability beyond its original training dataset.

%\vspace{-3pt}
\section{Conclusion}

In this paper, we propose a new task named source-free domain adaptive human pose estimation that places an emphasis on the privacy of source data. Additionally, we propose a new framework that includes source-protect and target-relevant modules, which aim to alleviate the issues of catastrophic forgetting of source and the sparsity of spatial distributions, respectively. Our approach is evaluated on hand and human pose datasets through extensive experiments, demonstrating that it outperforms state-of-the-art methods by a considerable margin.

\appendix

\section{Overview}

The supplementary material is organized into the following sections:

\begin{itemize}%[noitemsep,leftmargin=*] 
    \item Section \ref{sec:qua}: Additional qualitative results on FreiHand and Human3.6M datasets.
    \item Section \ref{sec:ab-fr}: Additional ablation of framework on FreiHand and LSP datasets.
    \item Section \ref{sec:ab-loss}: Additional ablation of losses on FreiHand and LSP datasets. 
    \item Section \ref{sec:gen}: Domain generalization to unseen domains based on models trained on domain adaptation tasks.

\end{itemize}

\section{Additional Qualitative Results}
\label{sec:qua}

\begin{figure}[!ht]
  \centering
   \includegraphics[width=0.99\linewidth]{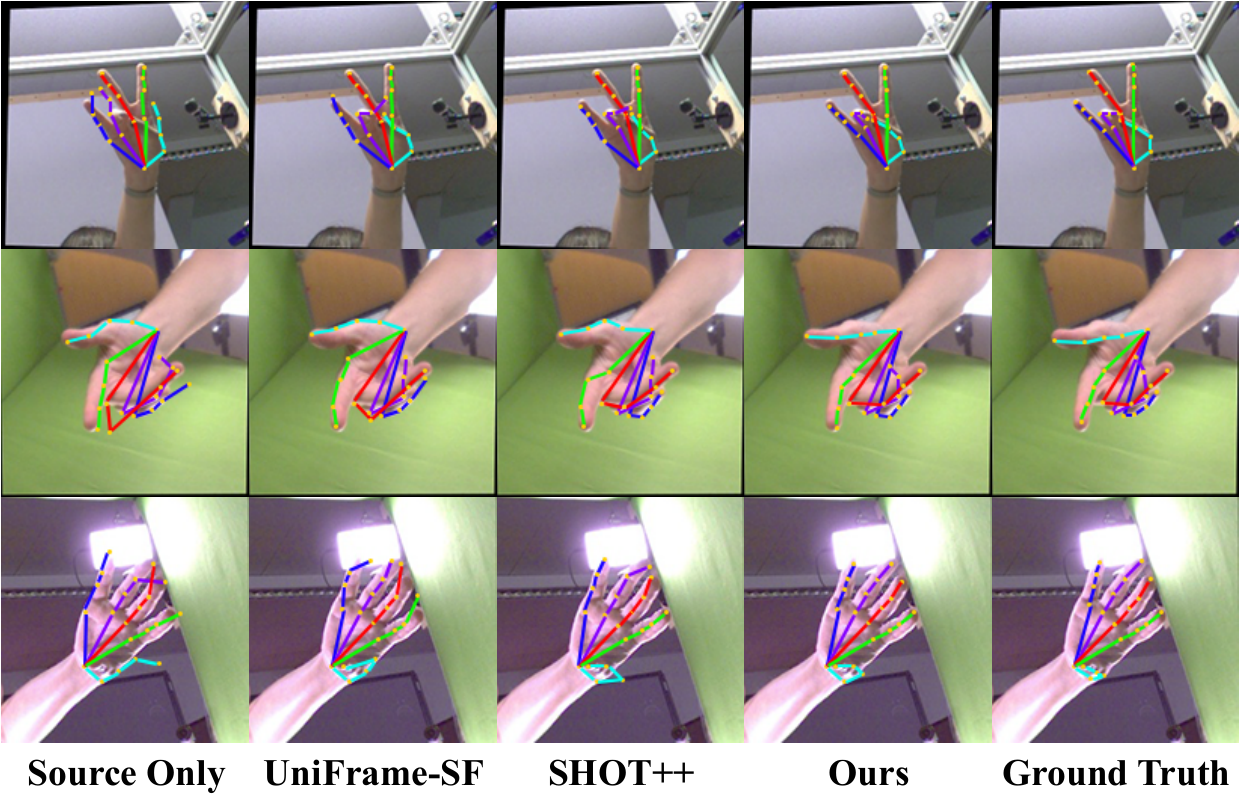}

   \caption{ Qualitative results on FreiHand dataset (Best view with zoom in)
   }
   \label{fig:freihand}
\end{figure}

\begin{figure}[!ht]
  \centering
   \includegraphics[width=0.99\linewidth]{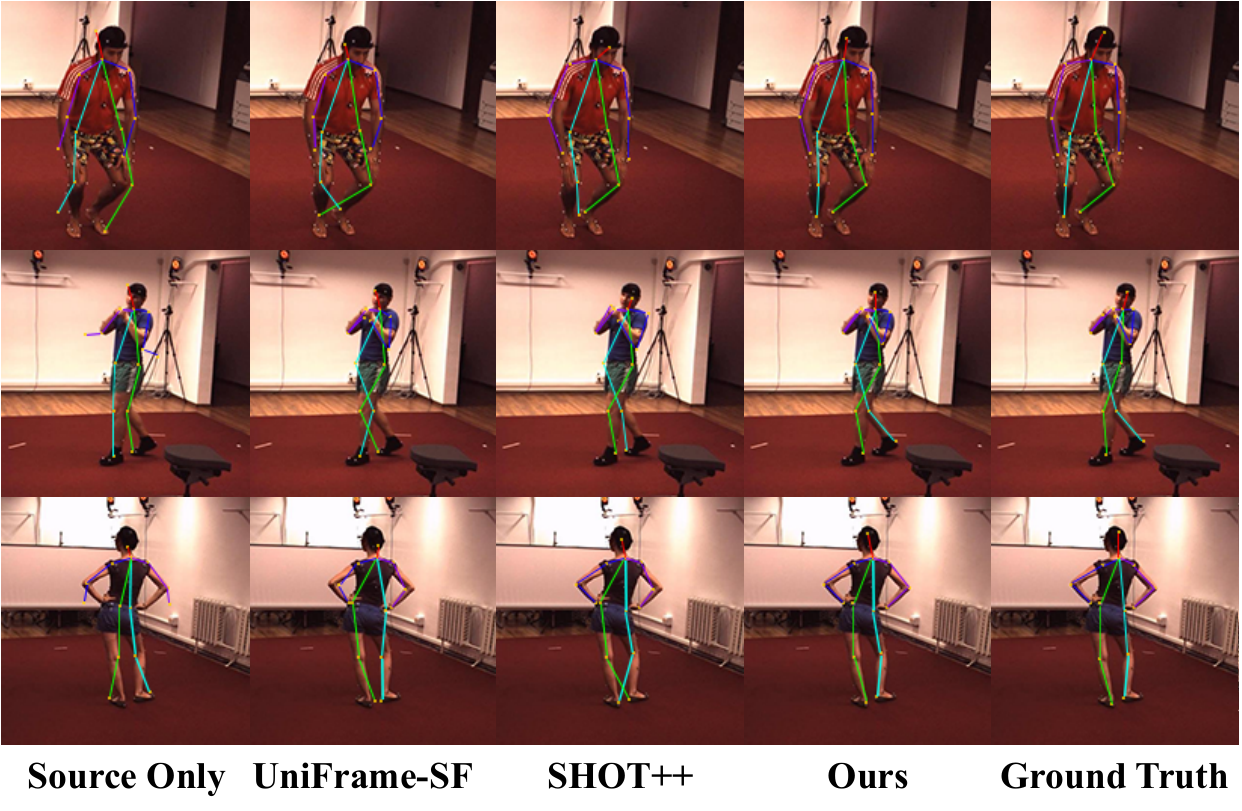}

   \caption{ Qualitative results on Human3.6M dataset (Best view with zoom in)
   }
   \label{fig:h36m}
\end{figure}

In this section, we offer additional qualitative results on the RHD$\rightarrow$FreiHand task and the SURREAL$\rightarrow$Human3.6M task. Results are exhibited in Fig. \ref{fig:freihand} and Fig. \ref{fig:h36m}.% There are four methods listed in these figures as \textbf{Source Only},  \textbf{UniFrame-SF} \cite{kim2022unified}, \textbf{SHOT++} \cite{liang2022shot++}, \textbf{Ours}, and the \textbf{Ground Truth}. 
These figures show that our method outperforms other competing methods, predicting more accurate poses in the target domain.

\section{Additional Ablation of Framework}
\label{sec:ab-fr}

Our method contains two modules in Step A \& Step B as the source-protect module (\textbf{SP}) and the target-relevant module (\textbf{TR}) separately, and here we focus on their functions.  Mutual Mean Teaching (\textbf{MMT}) \cite{ge2020mmt} is a domain adaptation strategy that preserves source information. We utilize MMT as the baseline method to evaluate the effectiveness of TR. Furthermore, comparing MMT with SP enables us to determine if our SP module outperforms MMT. Table \ref{tab:rf-ab} and \ref{tab:sl-ab} show the ablation study of frameworks on RHD$\rightarrow$ FreiHand and SURREAL$\rightarrow$ LSP with different combinations of these modules.

\begin{table}[!ht]
    \scriptsize
    \centering
    \caption{Ablation of Frameworks on RHD $\rightarrow$ FreiHand}
    \resizebox{0.9\linewidth}{!}{%
    \begin{tabular}{ccccccccc}
          \toprule
          Method &  MCP &  PIP  &  DIP & Fin & All \\
         \hline
         %{PM+PS} &  78.9 & 70.4 & 71.9 & 65.2 & 74.8\\
         {MMT \cite{ge2020mmt}} & 39.6 & 60.4 & 60.0 & 57.8 & 52.6 \\
         {MMT \cite{ge2020mmt} +TR} & 41.5 & 62.1 & 63.9 & 60.4 & 55.9\\
         {SP+TR (Ours)} & 43.7 & 65.9 & 66.6 & 63.1 & 58.8\\
         \bottomrule
    \end{tabular}%
    }
    %\vspace{-5pt}
\label{tab:rf-ab}
\end{table}

%\vspace{-8pt}
\begin{table}[!ht]
    \scriptsize
    \centering
    \caption{Ablation of Frameworks on SURREAL $\rightarrow$ LSP}
    \resizebox{0.97\linewidth}{!}{%
    \begin{tabular}{ccccccccc}
          \toprule
          Method &  Sld & Elb & Wrist & Hip & Knee & Ankle & All \\
         \hline
         %{PM+PS} & 74.5 & 84.0 & 72.7 & 47.6 & 81.5 & 83.2 & 74.4\\
         {MMT \cite{ge2020mmt}} & 60.9 & 70.9 & 70.3 & 81.1 & 79.3 & 72.8 & 71.5\\
         {MMT \cite{ge2020mmt} +TR} & 65.2 & 79.6 & 81.4 & 82.3 & 82.8 & 79.7 & 77.1\\ 
         {SP+TR (Ours)} & 70.7 & 85.4 & 83.8 & 86.6 & 85.2 & 85.0 & 83.2\\   
         \bottomrule
    \end{tabular}%
    }
    %\vspace{-10pt}
\label{tab:sl-ab}
\end{table}

%In this section, we offer additional ablations of the framework on the other two tasks and the results are shown in Table \ref{tab:rf-ab} and \ref{tab:sl-ab}.

The results clearly demonstrate that both SP and TR contribute to improving the model's performance. Specifically, TR enhances the model's accuracy by $3.3\%$ on RHD $\rightarrow$ FreiHand and $5.6\%$ on SURREAL $\rightarrow$ LSP, while SP leads to an improvement of $2.9\%$ on RHD $\rightarrow$ FreiHand and $5.1\%$ on SURREAL $\rightarrow$ LSP compared to MMT. Notably, the two proposed modules provide similar levels of improvement.

\section{Additional Ablation of Losses}
\label{sec:ab-loss}

We performed a detailed ablation study on the three proposed losses, namely $\mathcal{L}_{res}$, $\mathcal{L}_{cst}$, and $\mathcal{L}_{im}$, using the RHD$\rightarrow$ FreiHand and SURREAL$\rightarrow$ LSP tasks. Tables \ref{tab:rf-ab-ls} and \ref{tab:sl-ab-ls} present the results. 

%\vspace{-7pt}
\begin{table}[!ht]
    \scriptsize
    \centering
    \caption{Ablation of Losses on RHD $\rightarrow$ FreiHand}
    \resizebox{0.97\linewidth}{!}{%
    \begin{tabular}{ccccccccc}
          \toprule
          Method &  MCP &  PIP  &  DIP & Fin & All \\
         \hline
         {Baseline} & 41.2 & 63.5 & 63.8 & 60.9 & 56.3\\
         {$\mathcal{L}_{res}$} & 41.6 & 64.0 & 64.4 & 61.5 & 56.9\\
         {$\mathcal{L}_{cst}$} & 41.8 & 64.7 & 65.0 & 62.2 & 57.9\\
         {$\mathcal{L}_{im}$} & 41.5 & 64.4 & 64.8 & 61.7 & 57.1\\
         {$\mathcal{L}_{cst} \& \mathcal{L}_{im}$} & 42.3 & 65.0 & 65.4 & 62.5 & 58.1\\
         {$\mathcal{L}_{res} \& \mathcal{L}_{cst} \& \mathcal{L}_{im}$} & 43.7 & 65.9 & 66.6 & 63.1 & 58.8\\
         \bottomrule
    \end{tabular}%
    }
    %\vspace{-10pt}
\label{tab:rf-ab-ls}
\end{table}

%\vspace{-5pt}
\begin{table}[!ht]
    \scriptsize
    \centering
    \caption{Ablation of Losses on SURREAL $\rightarrow$ LSP}
    \resizebox{0.97\linewidth}{!}{
    \begin{tabular}{ccccccccc}
          \toprule
          Method &  Sld & Elb & Wrist & Hip & Knee & Ankle & All \\
         \hline
         {Baseline} & 69.9 & 82.1 & 81.3 & 84.5 & 82.7 & 80.4 & 80.3\\
         {$\mathcal{L}_{res}$} & 70.2 & 82.8 & 81.6 & 85.0 & 83.2 & 80.5 & 80.9\\
         {$\mathcal{L}_{cst}$} & 70.5 & 83.5 & 82.0 & 85.8 & 84.0 & 83.7 & 82.0\\
         {$\mathcal{L}_{im}$} & 70.3 & 83.0 & 81.9 & 85.2 & 83.4 & 82.2 & 81.3\\
         {$\mathcal{L}_{cst} \& \mathcal{L}_{im}$} & 70.6 & 84.8 & 82.7 & 86.0 & 84.6 & 84.1 & 82.5\\
         {$\mathcal{L}_{res} \& \mathcal{L}_{cst} \& \mathcal{L}_{im}$} & 70.7 & 85.4 & 83.8 & 86.6 & 85.2 & 85.0 & 83.2\\  
         \bottomrule
    \end{tabular}%
    }
\label{tab:sl-ab-ls}
\end{table}

%In this section, we offer additional ablations of the framework on the other two tasks and the results are shown in Table \ref{tab:rf-ab-ls} and \ref{tab:sl-ab-ls}. 

We observe that each loss is able to boost the model's performance. Simply applying $\mathcal{L}_{res}$ leads to an increase of $0.6\%$ in RHD$\rightarrow$ FreiHand and SURREAL$\rightarrow$ LSP. $\mathcal{L}_{cst}$ causes an improvement of $1.6\%$ in RHD$\rightarrow$ FreiHand and $1.7\%$ in SURREAL$\rightarrow$ LSP. As for $\mathcal{L}_{im}$, adding it achieves a performance gain of $0.8\%$ in RHD$\rightarrow$ FreiHand and $1.2\%$ in SURREAL$\rightarrow$ LSP. In addition, we observe that $\mathcal{L}_{cst}$ has a more significant impact than $\mathcal{L}_{res}$ or $\mathcal{L}_{im}$, as it yields greater improvements when compared to the other two.% losses.

\section{Generalization to Unseen Domains}
\label{sec:gen}

Following prior works \cite{li2021synthetic,kim2022unified}, we also conduct experiments on the generalization to unseen domains. For hand pose estimation, we use models adapted in the RHD$\rightarrow$H3D task and evaluate their performances on the validation set of FreiHand, as shown in Table \ref{tab:frei_gene}. For human pose estimation, we use models adapted in the SURREAL$\rightarrow$LSP task and evaluate their performance on Human3.6M, as shown in Table \ref{tab:h36m_gene}.

\begin{table}[!ht]
    \scriptsize
    \centering
    \caption{Domain Generalization on FreiHand}
    \resizebox{0.9\linewidth}{!}{%
    \begin{tabular}{rcccccccc}
          \toprule
          Method & SF &  MCP &  PIP  &  DIP & Fin & All \\
         \hline
         {Source-only} & - & 34.9 & 48.7 & 52.4 & 48.5 & 45.8\\
         %{Oracle} & 92.8 & 90.3 & 87.7 & 78.5 & 87.2\\
         \hline
         CC-SSL \cite{mu2020learning} (CVPR'20) & $\times$ & 34.3 & 46.3 & 48.4 & 44.4 & 42.6 \\
         MDAM \cite{li2021synthetic} (CVPR'21) & $\times$ & 29.6 & 46.6 & 50.0 & 45.3 & 42.2\\
         RegDA \cite{jiang2021regressive} (CVPR'21) & $\times$ & 37.8 & 51.8 & 53.2 & 47.5 & 46.9\\
         UniFrame \cite{kim2022unified} (ECCV'22) & $\times$ & 35.6 & 52.3 & 55.4 & 50.6 & 47.1\\
         \hline
         RegDA-SF \cite{jiang2021regressive} (CVPR'21) & \checkmark &  30.5 & 47.6 & 50.6 & 44.9 & 42.5\\         
         SHOT \cite{liang2020we} (ICML'20) & \checkmark &  32.0 & 48.1 & 49.9 & 42.4 & 41.8\\
         UniFrame-SF \cite{kim2022unified} (ECCV'22) & \checkmark & 32.7 & 48.5 & 51.3 & 45.7 & 43.0\\
         SHOT++ \cite{liang2022shot++} (TPAMI'22) & \checkmark & 33.6 & 49.2 & 52.5 & 47.0 & 44.6 \\
         {Ours} & \checkmark &  \textbf{34.4} & \textbf{50.8} & \textbf{54.7} & \textbf{48.3} & \textbf{46.2}\\    
                 
         \bottomrule
    \end{tabular}%
    }
    \vspace{-2pt}
\label{tab:frei_gene}
\end{table}

\begin{table}[!ht]
    \scriptsize
    \centering
    \caption{Domain Generalization on Human3.6M}
    \resizebox{0.97\linewidth}{!}{%
    \begin{tabular}{rcccccccc}
          \toprule
          Method & SF &  Sld & Elb & Wrist & Hip & Knee & Ankle & All \\
         \hline
         {Source-only} & - & 51.5 & 65.0 & 62.9 & 68.0 & 68.7 & 67.4 & 63.9\\
         %{Oracle} & 95.3 & 91.8 & 86.9 & 95.6 & 94.1 & 93.6 & 92.9\\
         \hline
         CC-SSL \cite{mu2020learning} (CVPR'20) & $\times$ & 52.7 & 76.9 & 63.1 & 31.6 & 75.7 & 72.9 & 62.2 \\
         MDAM \cite{li2021synthetic} (CVPR'21) & $\times$ & 54.4 & 75.3 & 62.1 & 21.6 & 70.4 & 69.2 & 58.8\\
         RegDA \cite{jiang2021regressive} (CVPR'21) & $\times$ & 76.9 & 80.2 & 69.7 & 52.0 & 80.3 & 80.0 & 73.2\\
         UniFrame \cite{kim2022unified} (ECCV'22) & $\times$ & 77.0 & 85.9 & 73.8 & 47.6 & 80.7 & 80.6 & 74.3\\
         \hline
         RegDA-SF \cite{jiang2021regressive} (CVPR'21) & \checkmark &  67.4 & 74.1 & 65.8 & 47.4 & 71.8 & 74.0 & 65.6\\         
         SHOT \cite{liang2020we} (ICML'20) & \checkmark &  68.6 & 75.8 & 67.0 & 48.1 & 72.4 & 74.4 & 66.2\\
         UniFrame-SF \cite{kim2022unified} (ECCV'22) & \checkmark & 68.4 & 74.7 & 66.0 & 48.3 & 72.2 & 74.9 & 66.6\\
         SHOT++ \cite{liang2022shot++} (TPAMI'22) & \checkmark & 69.7 & 76.0 & 66.4 & \textbf{48.8} & 73.4 & 75.8 & 67.9\\
         {Ours} & \checkmark &  \textbf{73.6} & \textbf{79.8} & \textbf{68.3} & 48.0 & \textbf{75.9} & \textbf{77.7} & \textbf{70.5}\\    
         \bottomrule
    \end{tabular}
    }
    \vspace{-2pt}
\label{tab:h36m_gene}
\end{table}

From these two tables, we can see that our model outperforms the second-best source-free approach for a lead of $1.6\%$ on FreiHand and $2.6\%$ on Human3.6M. %, which is not as large as that of the adaptation tasks presented in the main paper. This is partly because our model is designed for adaptation problems, not for generalization settings specifically. Nevertheless, our method still achieves the best results in these generalization experiments as compared to the state-of-the-art, demonstrating its superiority.

{\small
\bibliographystyle{ieee_fullname}
\bibliography{egbib}

\begin{thebibliography}{10}\itemsep=-1pt

\bibitem{chen2022contrastive}
Dian Chen, Dequan Wang, Trevor Darrell, and Sayna Ebrahimi.
\newblock Contrastive test-time adaptation.
\newblock In {\em Proceedings of the IEEE/CVF Conference on Computer Vision and
  Pattern Recognition (CVPR)}, pages 295--305, June 2022.

\bibitem{chen2020simple}
Ting Chen, Simon Kornblith, Mohammad Norouzi, and Geoffrey Hinton.
\newblock A simple framework for contrastive learning of visual
  representations.
\newblock In {\em International conference on machine learning}, pages
  1597--1607. PMLR, 2020.

\bibitem{chen2018cascaded}
Yilun Chen, Zhicheng Wang, Yuxiang Peng, Zhiqiang Zhang, Gang Yu, and Jian Sun.
\newblock Cascaded pyramid network for multi-person pose estimation.
\newblock In {\em Proceedings of the IEEE Conference on Computer Vision and
  Pattern Recognition}, pages 7103--7112, 2018.

\bibitem{deng2021cluster}
Wanxia Deng, Qing Liao, Lingjun Zhao, Deke Guo, Gangyao Kuang, Dewen Hu, and Li
  Liu.
\newblock Joint clustering and discriminative feature alignment for
  unsupervised domain adaptation.
\newblock {\em IEEE Transactions on Image Processing}, pages 7842--7855, 2021.

\bibitem{ganin2015unsupervised}
Yaroslav Ganin and Victor Lempitsky.
\newblock Unsupervised domain adaptation by backpropagation.
\newblock In {\em International Conference on Machine Learning}, pages
  1180--1189, 2015.

\bibitem{ge2020mmt}
Yixiao Ge, Dapeng Chen, and Hongsheng Li.
\newblock Mutual mean-teaching: Pseudo label refinery for unsupervised domain
  adaptation on person re-identification.
\newblock In {\em International Conference on Learning Representations}, 2020.

\bibitem{goodfellow2020generative}
Ian Goodfellow, Jean Pouget-Abadie, Mehdi Mirza, Bing Xu, David Warde-Farley,
  Sherjil Ozair, Aaron Courville, and Yoshua Bengio.
\newblock Generative adversarial networks.
\newblock {\em Communications of the ACM}, 63(11):139--144, 2020.

\bibitem{han2022transpar}
Zhongyi Han, Haoliang Sun, and Yilong Yin.
\newblock Learning transferable parameters for unsupervised domain adaptation.
\newblock {\em IEEE Transactions on Image Processing}, 31:6424--6439, 2022.

\bibitem{he2016deep}
Kaiming He, Xiangyu Zhang, Shaoqing Ren, and Jian Sun.
\newblock Deep residual learning for image recognition.
\newblock In {\em Proceedings of the IEEE Conference on Computer Vision and
  Pattern Recognition}, pages 770--778, 2016.

\bibitem{huang2017arbitrary}
Xun Huang and Serge Belongie.
\newblock Arbitrary style transfer in real-time with adaptive instance
  normalization.
\newblock In {\em Proceedings of the IEEE International Conference on Computer
  Vision}, pages 1501--1510, 2017.

\bibitem{ionescu2013human3}
Catalin Ionescu, Dragos Papava, Vlad Olaru, and Cristian Sminchisescu.
\newblock Human3. 6m: Large scale datasets and predictive methods for 3d human
  sensing in natural environments.
\newblock {\em IEEE Transactions on Pattern Analysis and Machine Intelligence},
  36(7):1325--1339, 2013.

\bibitem{jiang2021regressive}
Junguang Jiang, Yifei Ji, Ximei Wang, Yufeng Liu, Jianmin Wang, and Mingsheng
  Long.
\newblock Regressive domain adaptation for unsupervised keypoint detection.
\newblock In {\em Proceedings of the IEEE/CVF Conference on Computer Vision and
  Pattern Recognition}, pages 6780--6789, 2021.

\bibitem{jiang2022supervised}
Ruijie Jiang, Thuan Nguyen, Prakash Ishwar, and Shuchin Aeron.
\newblock Supervised contrastive learning with hard negative samples.
\newblock {\em arXiv preprint arXiv:2209.00078}, 2022.

\bibitem{jin2022branch}
Rui Jin, Jing Zhang, Jianyu Yang, and Dacheng Tao.
\newblock Multibranch adversarial regression for domain adaptative hand pose
  estimation.
\newblock {\em IEEE Transactions on Circuits and Systems for Video Technology},
  32(9):6125--6136, 2022.

\bibitem{johnson2010clustered}
Sam Johnson and Mark Everingham.
\newblock Clustered pose and nonlinear appearance models for human pose
  estimation.
\newblock In {\em British Machine Vision Conference}, volume~2, page~5.
  Citeseer, 2010.

\bibitem{kim2022unified}
Donghyun Kim, Kaihong Wang, Kate Saenko, Margrit Betke, and Stan Sclaroff.
\newblock A unified framework for domain adaptive pose estimation.
\newblock {\em arXiv preprint arXiv:2204.00172}, 2022.

\bibitem{kingma2014adam}
Diederik~P Kingma and Jimmy Ba.
\newblock Adam: A method for stochastic optimization.
\newblock {\em arXiv preprint arXiv:1412.6980}, 2014.

\bibitem{kurmi2021domain}
Vinod~K Kurmi, Venkatesh~K Subramanian, Vinay~P Namboodiri, {}, and {}.
\newblock Domain impression: A source data free domain adaptation method.
\newblock In {\em Proceedings of the IEEE/CVF Winter Conference on Applications
  of Computer Vision}, pages 615--625, 2021.

\bibitem{li2021synthetic}
Chen Li and Gim~Hee Lee.
\newblock From synthetic to real: Unsupervised domain adaptation for animal
  pose estimation.
\newblock In {\em Proceedings of the IEEE/CVF Conference on Computer Vision and
  Pattern Recognition}, pages 1482--1491, 2021.

\bibitem{li2020ma}
Rui Li, Qianfen Jiao, Wenming Cao, Hau-San Wong, and Si Wu.
\newblock Model adaptation: Unsupervised domain adaptation without source data.
\newblock In {\em 2020 IEEE/CVF Conference on Computer Vision and Pattern
  Recognition (CVPR)}, pages 9638--9647, 2020.

\bibitem{liang2020we}
Jian Liang, Dapeng Hu, and Jiashi Feng.
\newblock Do we really need to access the source data? source hypothesis
  transfer for unsupervised domain adaptation.
\newblock In {\em International Conference on Machine Learning}, pages
  6028--6039. PMLR, 2020.

\bibitem{liang2022shot++}
Jian Liang, Dapeng Hu, Yunbo Wang, Ran He, and Jiashi Feng.
\newblock Source data-absent unsupervised domain adaptation through hypothesis
  transfer and labeling transfer.
\newblock {\em IEEE Transactions on Pattern Analysis and Machine Intelligence},
  44(11):8602--8617, 2022.

\bibitem{lin2014microsoft}
Tsung-Yi Lin, Michael Maire, Serge Belongie, James Hays, Pietro Perona, Deva
  Ramanan, Piotr Doll{\'a}r, and C~Lawrence Zitnick.
\newblock Microsoft coco: Common objects in context.
\newblock In {\em European Conference on Computer Vision}, pages 740--755.
  Springer, 2014.

\bibitem{long2015learning}
Mingsheng Long, Yue Cao, Jianmin Wang, and Michael Jordan.
\newblock Learning transferable features with deep adaptation networks.
\newblock In {\em International Conference on Machine Learning}, pages 97--105,
  2015.

\bibitem{mu2020learning}
Jiteng Mu, Weichao Qiu, Gregory~D Hager, and Alan~L Yuille.
\newblock Learning from synthetic animals.
\newblock In {\em Proceedings of the IEEE/CVF Conference on Computer Vision and
  Pattern Recognition}, pages 12386--12395, 2020.

\bibitem{peng2022toward}
Qucheng Peng, Zhengming Ding, Lingjuan Lyu, Lichao Sun, and Chen Chen.
\newblock Toward better target representation for source-free and black-box
  domain adaptation.
\newblock {\em arXiv preprint arXiv:2208.10531}, 2022.

\bibitem{pinyoanuntapong2023gaitsada}
Ekkasit Pinyoanuntapong, Ayman Ali, Kalvik Jakkala, Pu Wang, Minwoo Lee,
  Qucheng Peng, Chen Chen, and Zhi Sun.
\newblock Gaitsada: Self-aligned domain adaptation for mmwave gait recognition.
\newblock {\em arXiv preprint arXiv:2301.13384}, 2023.

\bibitem{saito2018maximum}
Kuniaki Saito, Kohei Watanabe, Yoshitaka Ushiku, and Tatsuya Harada.
\newblock Maximum classifier discrepancy for unsupervised domain adaptation.
\newblock In {\em Proceedings of the IEEE Conference on Computer Vision and
  Pattern Recognition}, pages 3723--3732, 2018.

\bibitem{shi2022end}
Dahu Shi, Xing Wei, Liangqi Li, Ye Ren, and Wenming Tan.
\newblock End-to-end multi-person pose estimation with transformers.
\newblock In {\em Proceedings of the IEEE/CVF Conference on Computer Vision and
  Pattern Recognition}, pages 11069--11078, 2022.

\bibitem{sun2019deep}
Ke Sun, Bin Xiao, Dong Liu, and Jingdong Wang.
\newblock Deep high-resolution representation learning for human pose
  estimation.
\newblock In {\em Proceedings of the IEEE/CVF Conference on Computer Vision and
  Pattern Recognition}, pages 5693--5703, 2019.

\bibitem{tang2020unsupervised}
Hui Tang, Ke Chen, and Kui Jia.
\newblock Unsupervised domain adaptation via structurally regularized deep
  clustering.
\newblock In {\em Proceedings of the IEEE/CVF Conference on Computer Vision and
  Pattern Recognition}, pages 8725--8735, 2020.

\bibitem{tarvainen2017mean}
Antti Tarvainen and Harri Valpola.
\newblock Mean teachers are better role models: Weight-averaged consistency
  targets improve semi-supervised deep learning results.
\newblock {\em Advances in Neural Information Processing Systems}, 30, 2017.

\bibitem{tompson2015efficient}
Jonathan Tompson, Ross Goroshin, Arjun Jain, Yann LeCun, and Christoph Bregler.
\newblock Efficient object localization using convolutional networks.
\newblock In {\em Proceedings of the IEEE conference on computer vision and
  pattern recognition}, pages 648--656, 2015.

\bibitem{tompson2014joint}
Jonathan~J Tompson, Arjun Jain, Yann LeCun, and Christoph Bregler.
\newblock Joint training of a convolutional network and a graphical model for
  human pose estimation.
\newblock {\em Advances in neural information processing systems}, 27, 2014.

\bibitem{tzeng2014deep}
Eric Tzeng, Judy Hoffman, Ning Zhang, Kate Saenko, and Trevor Darrell.
\newblock Deep domain confusion: Maximizing for domain invariance.
\newblock {\em arXiv preprint arXiv:1412.3474}, 2014.

\bibitem{varol2017learning}
Gul Varol, Javier Romero, Xavier Martin, Naureen Mahmood, Michael~J Black, Ivan
  Laptev, and Cordelia Schmid.
\newblock Learning from synthetic humans.
\newblock In {\em Proceedings of the IEEE Conference on Computer Vision and
  Pattern Recognition}, pages 109--117, 2017.

\bibitem{wang2022regularizing}
Haixin Wang, Lu Zhou, Yingying Chen, Ming Tang, and Jinqiao Wang.
\newblock Regularizing vector embedding in bottom-up human pose estimation.
\newblock In {\em Computer Vision--ECCV 2022: 17th European Conference, Tel
  Aviv, Israel, October 23--27, 2022, Proceedings, Part VI}, pages 107--122.
  Springer, 2022.

\bibitem{xia2021adaptive}
Haifeng Xia, Handong Zhao, and Zhengming Ding.
\newblock Adaptive adversarial network for source-free domain adaptation.
\newblock In {\em Proceedings of the IEEE/CVF International Conference on
  Computer Vision}, pages 9010--9019, 2021.

\bibitem{xiao2018simple}
Bin Xiao, Haiping Wu, and Yichen Wei.
\newblock Simple baselines for human pose estimation and tracking.
\newblock In {\em Proceedings of the European Conference on Computer Vision
  (ECCV)}, pages 466--481, 2018.

\bibitem{zhang2020distribution}
Feng Zhang, Xiatian Zhu, Hanbin Dai, Mao Ye, and Ce Zhu.
\newblock Distribution-aware coordinate representation for human pose
  estimation.
\newblock In {\em Proceedings of the IEEE/CVF Conference on Computer Vision and
  Pattern Recognition}, pages 7093--7102, 2020.

\bibitem{zhao2020h3d}
Zhengyi Zhao, Tianyao Wang, Siyu Xia, and Yangang Wang.
\newblock Hand-3d-studio: A new multi-view system for 3d hand reconstruction.
\newblock In {\em ICASSP 2020 - 2020 IEEE International Conference on
  Acoustics, Speech and Signal Processing (ICASSP)}, pages 2478--2482, 2020.

\bibitem{zimmermann2017learning}
Christian Zimmermann and Thomas Brox.
\newblock Learning to estimate 3d hand pose from single rgb images.
\newblock In {\em Proceedings of the IEEE International Conference on Computer
  Vision}, pages 4903--4911, 2017.

\bibitem{zimmermann2019freihand}
Christian Zimmermann, Duygu Ceylan, Jimei Yang, Bryan Russell, Max Argus, and
  Thomas Brox.
\newblock Freihand: A dataset for markerless capture of hand pose and shape
  from single rgb images.
\newblock In {\em Proceedings of the IEEE/CVF International Conference on
  Computer Vision}, pages 813--822, 2019.

\end{thebibliography}
}

\end{document}